# Recurrent U-Net-Based Graph Neural Network (RUGNN) for Accurate Deformation Predictions in Sheet Material Forming

Yingxue Zhao[1], Qianyi Chen[2], Haoran Li[1], Haosu Zhou[1], Hamid Reza Attar[1], Tobias Pfaff[3], Tailin Wu[2], Nan Li[1*]

[1] Dyson School of Design Engineering, Imperial College London, London, UK

[2] School of Engineering, Westlake University, Hangzhou, China

[3] Google DeepMind, London, UK

[*]Corresponding author. E-mail address: n.li09@imperial.ac.uk (N. Li)

## Abstract

In recent years, various artificial intelligence-based surrogate models have been proposed to provide rapid manufacturability predictions of material forming processes. However, traditional AI-based surrogate models, typically built with scalar or image-based neural networks, are limited in their ability to capture complex 3D spatial relationships and to operate in a permutation-invariant manner. To overcome these issues, emerging graph-based surrogate models are developed using graph neural networks. This study developed a new graph neural network surrogate model named Recurrent U Net-based Graph Neural Network (RUGNN). The RUGNN model can achieve accurate predictions of sheet material deformation fields across multiple forming timesteps. The RUGNN model incorporates Gated Recurrent Units (GRUs) to model temporal dynamics and a U-Net inspired graph-based downsample/upsample mechanism to handle spatial long-range dependencies. A novel 'node-to-surface' contact representation method was proposed, offering significant improvements in computational efficiency for large-scale contact interactions. The RUGNN model was validated using a cold forming case study and a more complex hot forming case study using aluminium alloys. Results demonstrate that the RUGNN model provides accurate deformation predictions closely matching ground truth FE simulations and outperforming several baseline GNN architectures. Model tuning was also performed to identify suitable hyperparameters, training strategies, and input feature representations. These results demonstrate that RUGNN is a reliable approach to support sheet material forming design by enabling accurate manufacturability predictions.

## Keywords

Machine learning, graph neural networks, gated recurrent unit, surrogate models, design for manufacturing, contact modelling

# 1. Introduction

Sheet material forming involves non-linear, dynamic manufacturing processes that deform sheet materials into complex 3D components [1]. A typical process, stamp forming, uses a set of rigid-body tools to shape sheet material into the desired form. In sheet material stamping, tool design plays a crucial role in determining the outcomes of the stamping process [2]. The configuration of the tools influences the material flow and stress distribution, which affect the likelihood of manufacturing-induced defects such as cracking and wrinkling [3, 4]. These manufacturing-induced defects can compromise the structural integrity of the post-formed product. Traditionally, early-stage tool design primarily depends on feedback obtained from Finite Element (FE) simulations. The FE simulations provide results such as thinning and displacement fields, which help identify areas at risk of excessive thinning and cracking, or regions prone to wrinkling or folding. By analysing these FE simulation results, tool designers optimise tool geometries through a trial-and-error approach. However, as material forming problems become more complex and non-linear, the FE simulations can become increasingly time-consuming.

To overcome these limitations, various Machine Learning (ML)-based surrogate models have been trained in a data-driven manner to provide more rapid manufacturability feedback to guide design changes. For instance, Convolutional Neural Network (CNN) models have been extensively applied in material forming related domains such as predicting post-formed thinning fields and displacement fields [5-7], as well as predicting load transfer paths in plate structures [8]. However, image-based CNN models rely on 2D grid-based representations, which limit their flexibility in capturing complex spatial structures, particularly for 3D geometries represented by meshes or other irregular formats [9]. Additionally, it is challenging for image-based CNNs to represent complex spatial interactions, such as contact interactions between rigid and deformable objects in various of engineering applications. Recently, Graph Neural Networks (GNNs) have been introduced as a new type of models tailored to process graph-based data. In GNNs, the graph-based data typically consists of node features and their interconnection (edge) features [10, 11]. GNNs update graph-based data through multiple message-passing layers, enabling each graph node to aggregate and update information from its neighbours via the graph edges [10, 12]. In mechanics-related applications, mechanical structures can be converted into data formats such as FE meshes or point clouds that can be represented as graph-based data [13]. These graph-based representations provide greater flexibility in learning from 3D geometries with irregular topologies. Additionally, the graph construction method improves the flexibility for modelling spatial relationships between interacting objects, such as contact interactions. Furthermore, GNNs are typically designed to be permutation-invariant and to employ shared update functions across nodes and edges. This architectural design supports localised computations and allows the model to learn local rules that are more robust to geometric variations beyond the training set [11, 14].

Although GNNs have shown great potential in mechanics-related domains, their application remains in its early stages [15]. Most GNN studies focus on small graphs with a limited number of nodes and edges [15-17]. Scaling GNNs to large graphs used in real-world industrial applications require more message-passing layers to propagate information across longer spatial distances, which significantly increases computational memory requirements and processing time. Additionally, current GNN studies on mechanical systems often use simplified boundary conditions such as nodal forces [18] that can be directly included as graph node features. A few innovative GNN studies have designed novel data-representation method and architectures to model complex boundary conditions involving contact interactions [11, 19, 20]. These contact-modelling GNN models are mainly designed for computer vision or robotics applications with relatively small-scale, simple rigid bodies, like ball or pin shapes. For large-scale contact interactions with complex-shaped rigid bodies, these methods can become computationally expensive and require longer processing times. Moreover, a few GNN-based learned simulators have been proposed to predict spatio-temporal simulation results of mechanical systems [11, 13, 19, 21-23]. These models predict results over multiple timesteps in an autoregressive manner. However, they often exhibit high accumulated errors, especially at later timesteps [11, 23]. Furthermore, most existing GNN simulators simplify mechanics-related problems with time-dependent materials, such as linear elastic [18, 24], bilinear elastic-plastic [23], and hyperelastic materials [25, 26]. The ability of GNNs to model mechanics-related problems with more non-linear and time-independent material properties has not yet been fully explored.

In the sheet material forming domain, the material behaviours to be simulated by GNNs are more complex, compared to the aforementioned studies. The as-formed components typically have complex 3D geometries [27], requiring larger numbers of nodes and edges to accurately represent the geometries, which can be computationally expensive. The boundary conditions in sheet material forming involve large-scale contact interactions between complex-shaped rigid body tools and the deformable blank sheet [28]. Furthermore, high prediction accuracy is required across all process timesteps to obtain realistic manufacturability results to effectively guide design improvements. Finally, commonly used sheet materials, such as high-strength aluminium alloy deformed under warm or hot conditions, exhibit elastic-viscoplastic behaviour that is strongly dependent on temperature and strain rate, and is highly non-linear and time-dependent [29, 30].

To address the research challenges posed by sheet material forming applications, a new Recurrent U-Net-based Graph Neural Network (RUGNN) framework is proposed in this study. This RUGNN framework is designed to accurately and efficiently predict the evolution of material deformations under different rigid tool geometries. The main contributions of this work are outlined as follows:

- The proposed RUGNN framework integrates Gated Recurrent Units (GRUs) in the processor blocks, improving the model capacity to capture temporal dynamics and significantly reduce autoregressive accumulated errors. It also incorporates a U-Net-based graph downsample/upsample mechanism, significantly enhancing the model's ability to manage spatial long-range dependencies.
- A novel node-to-surface contact representation method is proposed to model large-scale contact interactions while reducing the computational burden.
- The effectiveness of the RUGNN framework is demonstrated through cold and hot material forming case studies using aluminium alloys. The RUGNN model outperforms several GNN baseline models.
- Model hyperparameters, training strategies, and input feature representations are further analysed to provide a comprehensive evaluation of model performance.

The structure of this paper is as follows: Section 2 reviews relevant GNN studies in various mechanics-related domains. Section 3 outlines the stamp forming case studies and dataset generation processes. Section 4 explains the graph representation and processing method used in the RUGNN framework, while Section 5 provides a detailed description of the RUGNN architecture. Section 6 covers the training strategies and loss function formulation. Section 7 evaluates the predictive performance of the RUGNN framework across various stamp forming conditions and tool geometries, with comparisons to several baseline models.

## 2. Relevant GNN models

### 2.1 Relevant GNN models for modelling material deformation behaviours

GNN models have been extensively used to model material deformation behaviours, particularly for continuum structures such as beams, plates, and flags [15]. In GNNs, graph representation for continuum structures is commonly constructed from FE mesh, where graph nodes correspond to mesh nodes and graph edges represent connectivity between them. Node features often include external nodal forces, nodal coordinates, material properties, and one-hot vectors for node types. Edge features typically include relative positions and Euclidean distances between adjacent nodes [11, 18, 26, 31].

Several GNN-based frameworks were proposed to address challenges in deformation modelling across a range of physical scenarios. Pfaff et al. [11] introduced the MeshGraphNet (MGN), which simulates 3D cloth deformation under contact with a moving ball and quasi-static deformations of hyperelastic plates. Building on the success of MGN, various efforts were made to further enhance the framework for broader applications. Among them, a recurrent MGN-based model named Physics-informed edge recurrent simulator (Piers) extendes the research scope from quasi-static material deformations to dynamic responses in continuum deformable bodies [23]. This method models deformation as a sequence of graph updates. It also incorporates prior physical knowledge into hidden states and employs physics-informed loss functions. The Piers framework demonstrates improved accuracy in capturing complex and dynamic material behaviours during autoregressive predictions.

Several graph-based spatial downsample techniques were also proposed to reduce computational burden in processing large graphs. For example, Fortunato et al. [32] introduced Multiscale MGN. The model constructs inter-level edges to connect nodes between finer and coarser graphs, enabling message passing between them. Similarly, Deshpande et al. [31] proposed a U-Net-based Graph Convolutional Networks (GCN) to predict the displacement fields of 2D and 3D beam structures. This model uses input-independent weighted multi-channel aggregation layers, improving displacement field predictions.

### 2.2 Relevant GNN models for modelling contact between rigid and deformable bodies

External loads in FE simulations are not always explicitly included in graph inputs. Instead, they can be implicitly represented through contact interactions between objects. While most current GNN architectures handle simplified loading conditions such as point loads or uniformly distributed loads, few address complex contact-driven deformations. One notable contribution is the MGN, a multigraph GNN architecture capable of learning mesh-based self-collisions and object-to-object contact [11]. MGN introduces contact edges, generated using a predefined search radius to represent contact interactions. These edges propagate contact information and update node features in the deformable body. To improve contact modelling accuracy, Allen et al.[19] replace node-to-node contact edges with face-to-face contact edges, linking the sender and receiver surfaces of the rigid body. This modification leads to improvements in both predictive accuracy and computational efficiency. Saleh, et al. [20] incorporate positional information from both deformable and rigid objects and apply multi-head cross-attention mechanisms to model interactions. In this method, physics-based positional encoder is applied to the deformable and rigid objects, and multi-head attention captures interactions between their respective features [20].

However, these methods are primarily developed for modelling small-scale contact problems in computer vision and robotics, where the rigid bodies are relatively small, such as balls or pins. When applied to material forming, which typically involves larger contact areas, they become computationally expensive. In material forming, the rigid bodies (i.e., forming tools) are comparable in size to the deformed material, resulting in an excessive number of contact edges due to the large contact area. Processing these contact edges can be highly computationally demanding. A recent study by Rubanova, et al. [33] addresses this challenge by implicitly representing objects using Signed Distance Functions (SDFs). This method computes contact edges based on learned distance fields, which helps reduced the number of contact edges and the associated computational cost for large-scale contact interactions. However, each new rigid body requires retraining a separate SDF model, limiting its adaptability in settings with varying rigid tool shapes. To address these challenges in material forming applications, a node-to-surface contact modelling method is proposed. This method maps contact information directly onto the deformable object's nodes, offering a lightweight approach to model contact interactions. The details of this method are explained in Section 4.2.

### 2.3 Relevant spatio-temporal GNN models

Spatio-temporal surrogate models are widely used in domains such as traffic predictions or computer vision. For example, Yu et al.[34] introduced the Spatio-Temporal U-Network (ST-UNet) for graph-based time-series modelling. ST-UNet integrates spatial graph pooling with temporal dilated skip connections, enabling more accurate rollout predictions. In the mechanical domain, spatio-temporal surrogate modelling is an emerging area of research. For instance, MGN predicts the spatio-temporal evolution of fluid dynamics and structural deformations [11]. Building on this, Piers embeds Gated Recurrent Units (GRUs) into the edge update block of MGN. Compared to MGN, Piers improved the autoregressive prediction accuracy in modelling the spatio-temporal evolution of beam deformations and flag dynamics [23].

However, current GNN models face significant challenges in material forming applications. Existing models mainly focus on small-scale graphs, where global message passing is efficient with fewer message passing layers [23]. In contrast, sheet material forming requires larger-scale graphs to represent complex 3D geometries, which necessitate longer message-passing distances. This demands more message passing layers, which significantly increases computational costs and reduces training efficiency. Additionally, most existing GNN models have been evaluated using simplified material representations, such as linear elastic, bilinear elastic-plastic, or hyper-elastic materials. These materials are typically assumed to be time-independent. However, materials in forming processes at elevated temperatures often exhibit time-dependent behaviours, introducing additional nonlinearities.

To overcome these limitations, the RUGNN model integrates GRUs with a hierarchical graph downsample/upsample mechanism. GRUs enhance the model's ability to capture nonlinear temporal dependencies by handling edge features and incorporating them into node updates. Meanwhile, the hierarchical graph downsample/upsample mechanism facilitates efficient spatial long-range message passing. This combination enables the model to efficiently handle large-scale graphs with complex geometries and supports robust learning of nonlinear spatio-temporal dynamics

## 3. Problem statement and dataset generation

The primary task of this study is to develop a forward prediction model that simulates the aluminium forming process as a sequence of multiple timestep results. Given the initial mesh configurations of both blank and forming tools, the model predicts the deformation fields of the blank at subsequent timesteps throughout the forming process. This study focuses on the aluminium metal stamping, a forming process that shapes sheet aluminium materials into desired geometries. Datasets for the dome-shaped cold forming and bulkhead-shaped hot forming case studies were developed. The detailed simulation setup and data generation procedures are described in the following section.

### 3.1 Case study 1: Dome-shaped cold forming case study

Dome-shaped geometry is characterised by a double-curvature surface, which create a more complex stress and deformation state than a single-curvature geometry. This makes dome-shaped geometry an important baseline for evaluating material formability and tool design effectiveness [35]. The cold stamping simulation setup for dome-shaped case study is illustrated in Fig. 1 (a). The forming tools, including die, punch, blank holder, and spacer, are treated as rigid body objects. The blank is the deformable object. The die is fixed in all degrees of freedom, while the punch moves vertically at a prescribed stamping speed to push the blank into the die. The stamping speed is 500 mm/s. A blank holding force of 200 kN is applied to blank holders to support the blank edges, preventing excessive wrinkling. Spacers with a thickness of 2.2 mm are used to create a small offset between the die and blank holder to facilitate material flow. A simplified isotropic material model for coldwork Al2008 provided by PAMSTAMP material database is used. The initial blank thickness is set to 2 mm, and the friction coefficient is set to 0.12.

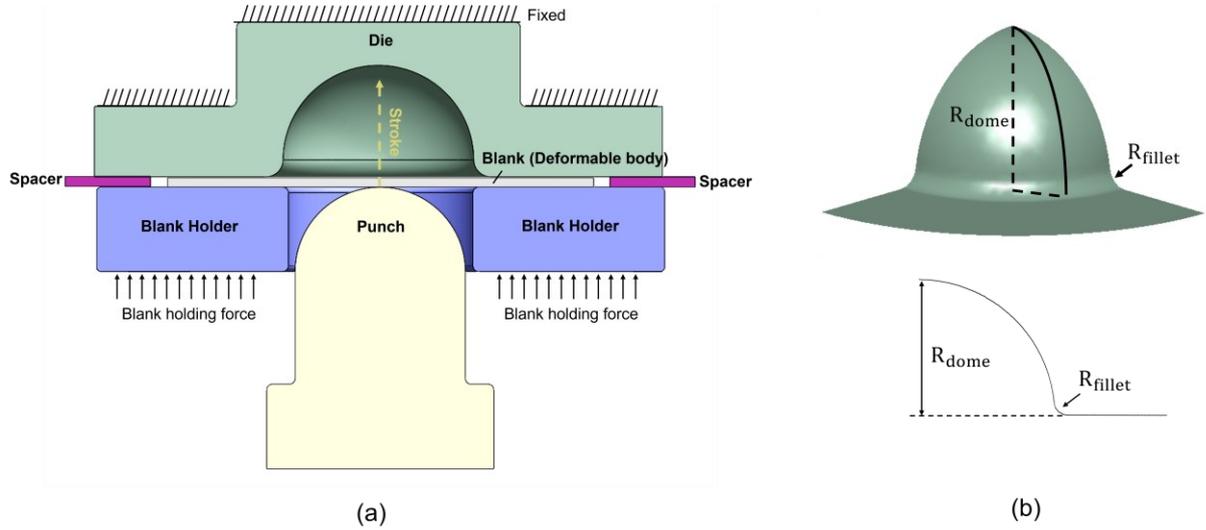

Fig. 1 (a) Sectional view of simulation setup for dome-shaped case study; (b) 3D and 2D sectional views of the dome-shaped tool design.

Different dome-shaped tool geometries were generated according to the Design of Experiment (DoE) setup shown in Fig. 1 (b). The die geometry is defined by $R_{fillet}$ and $R_{dome}$, with ranges of 60mm ≤ $R_{dome}$ ≤ 120mm, 10mm ≤ $R_{fillet}$ ≤ 20mm, respectively. The punch is designed with a 2 mm offset relative to the die. Notably, due to the symmetry of the dome geometry, only one quarter of the dome is modelled to reduce the computational cost of FE simulations. Symmetry planes is defined as boundary conditions. A uniform quarter-circle shape blank is used across all simulation case studies as shown in Fig. 5 (b).

An automated dataset generation pipeline was developed following the similar procedure in [5]. Tool geometry parameters were generated using Latin Hypercube (LHC) sampling within the previously defined DoE ranges. LHC sampling was performed separately for the training, validation, and testing sets to ensure comprehensive coverage of the design space. CAD models of the tool design variants were automatically generated in SolidWorks, meshed in HyperMesh, and simulated in PAMSTAMP. The final dataset consisted of 50 train samples, 20 validation samples, and 20 test samples. Each sample was simulated with a unique set of tool geometries and contained stamping results across 11 timesteps, capturing material deformation under various tool configurations.

### 3.2 Case study 2: Bulkhead-shaped hot forming case study

A more complex bulkhead-shaped hot stamping case study was conducted. Compared to the dome-shaped case, it involved larger-scale simulations and more complex geometry. The stamping simulations were performed under elevated temperatures, using a time-independent elastic-viscoplastic material model. The hot stamping simulation setup for the bulkhead-shaped case study is illustrated in Fig. 2 (a). The forming tools, including die, punch, blank holder, and spacer, are treated as rigid body obejcts. The blank is the deformable object. The die is fixed in all degrees of freedom, while the punch moves vertically under imposed displacement boundary conditions to deform the blank over the die. The blank holding force of 200 kN is applied to blank holders to support the blank edges, preventing excessive wrinkling. Spacers with thickness of 2.2 mm creates a small offset between the die and blank holder, which helps preventing rapid cooling at the blank's edges and facilitates material flow. The blank material is a temperature- and strain rate-dependent viscoplastic model for AA6082. This material model is originally adapted from Mohamed et al.[36] and incorporated into a material database in a previous study [5]. The initial blank thickness is set to 2 mm. The initial tooling temperature is set to 25 °C, the initial blank temperature is set to 500 °C, and friction coefficient is set to 0.1 [37]. The stamping speed is 500 mm/s.

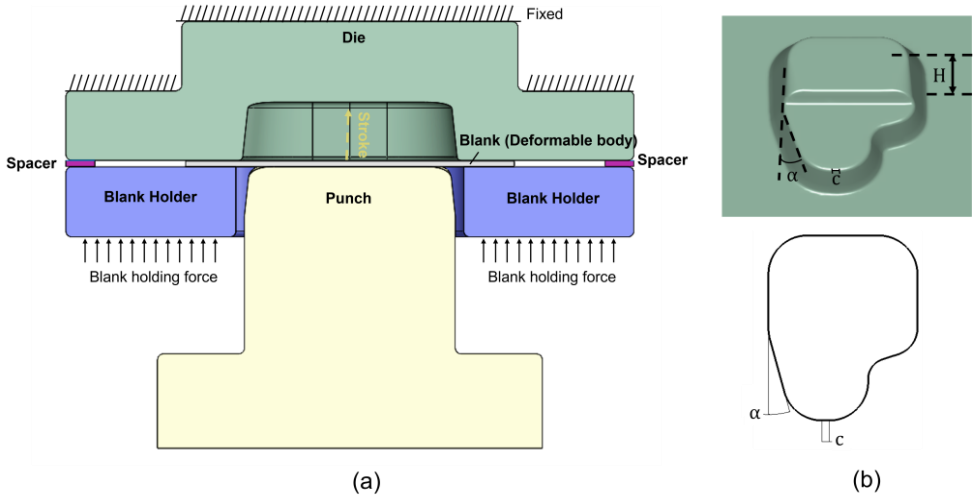

Fig. 2 (a) 3D and 2D plan views of the bulkhead-shaped tool design; (b) Sectional view of simulation setup for bulkhead-shaped case study.

Different bulkhead-shaped forming tools were generated using the DoE setup in Fig. 2 (b), where the 2D drawing illustrates the plan view of the top surface of the die. The die geometry is defined by the total height of the bulkhead H, the horizontal length c, and the inclination angle α, with ranges of 70mm ≤ H ≤ 100mm, 10 mm ≤ c ≤ 30mm, and 5 degrees ≤ α ≤ 20 degrees, respectively. A uniform blank shape, as shown in Fig. 5 (b), is used across all simulation case studies. The same automated dataset generation pipeline was used to generate different bulkhead-shaped dataset. The bulkhead-shaped case study did not perform hyperparameter tuning, validation set was therefore not omitted. A total of 100 train samples and 45 test samples were generated. Each sample was simulated with a unique set of tool geometries and contained stamping results across 11 timesteps.

## 4. Graph representation and processing

The RUGNN model operates on a sequence of graphs, each corresponding to a specific timestep in the stamping simulation. For both the dome-shaped and bulkhead-shaped case studies, the simulation is discretised into $T = 10$ timestep intervals, resulting in $T + 1 = 11$ timesteps from the undeformed initial blank to the fully formed final shape, with each timestep represented by a unique graph representation. In this section, we describe the graph representation method and data processing methods used in the proposed RUGNN model. Section 4.1 outlines the input and output feature vectors defined on the graph at each timestep. Section 4.2 introduces a contact modelling method that computes contact-related features based on the relative geometry between the deformable object and rigid tools. Section 4.3 presents a graph coarsening method to reduce the computational cost of message passing. It also explains how inter-level edges are constructed between the finer and coarser graphs to support message propagation across different graph resolutions.

### 4.1 Graph input and output features

According to Fig. 3, the graph structure is constructed from the FE mesh, where mesh nodes correspond to graph nodes and mesh edges correspond to graph edges. The graph representation at timestep t is denoted as $G^t = (\boldsymbol{u}, V^t, E^t)$. $V^t$ is the set of graph nodes. Each node in $V^t$ is associated with input node feature vectors $\boldsymbol{v}_n^t$, and $\boldsymbol{v}_{ncont}^t$, and an output node feature vector $\Delta \boldsymbol{x}_n^{t+1}$. $E^t$ is the set of graph edges. Each edge is associated with an edge feature vector $\boldsymbol{e}_e^t$. $\boldsymbol{u}$ is the global feature vector that encodes timestep-wise information shared across the graph. The detailed description of input and output feature vectors within each graph in timestep t are summarised in Table 1.

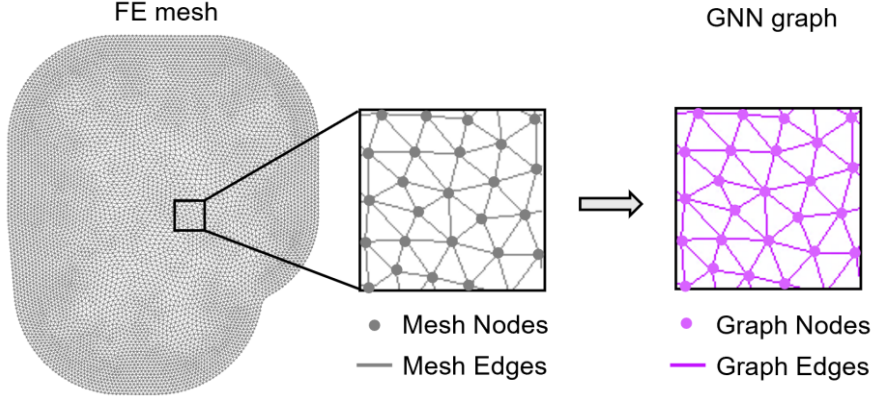

Fig. 3 Conversion from the FE mesh to the GNN graph structure.

**Table 1** Graph input and output feature vectors for two case studies.

| Case study | Node ($v_n^t$) | Edge ($e_e^t$) | Global ($u$) | Contact ($v_{ncont}^t$) | Output ($\Delta x_n^{t+1}$) |
|---|---|---|---|---|---|
| Dome | $\Delta x_n^t = x_n^t - x_n^{t-1}$; $\boldsymbol{b} = (b_x, b_y, b_z)$ | $e_{ij}^0 = x_{n_i}^0 - x_{n_j}^0;\ \|e_{ij}^0\|_2$ <br> $e_{ij}^t = x_{n_i}^t - x_{n_j}^t;\ \|e_{ij}^t\|_2$ | $\Delta t$, stroke | $\frac{1}{d_n^t};\ \boldsymbol{n}^t$ | $x_n^{t+1} - x_n^t$ |
| Bulkhead | $\Delta x_n^t = x_n^t - x_n^{t-1}$ | $e_{ij}^0 = x_{n_i}^0 - x_{n_j}^0;\ \|e_{ij}^0\|_2$ <br> $e_{ij}^t = x_{n_i}^t - x_{n_j}^t;\ \|e_{ij}^t\|_2$ | $\Delta t$, stroke | $\frac{1}{d_n^t};\ \boldsymbol{n}^t$ | $x_n^{t+1} - x_n^t$ |

According to Table 1, at each timestep t, the feature vector $v_n^t$ of an arbitrary node include the relative position difference $\Delta x_n^t$ between the node's current position $x_n^t$ and its position in the previous timestep $x_n^{t-1}$. At the initial timestep (t = 0), zero paddings are used as $\Delta x_n^t$. In the dome-shaped case study with symmetric boundary conditions, a one-hot vector $\boldsymbol{b}$ is used to distinguish fixed boundary node. The components $b_x$, $b_y$, and $b_z$ represents the fixed boundary condition in x, y, and z directions, respectively. Each component is 0 or 1 depending on whether the node is fixed or unfixed in that direction.

Edge feature vector $e_e^t$ contains features calculated from the undeformed geometry in the initial timestep, and the deformed geometry in current timestep. Specifically, $e_{ij}^0$ and $\|e_{ij}^0\|_2$ represent the relative distances and the Euclidean distance between source node i and target node j at the initial timestep (t = 0), respectively. $e_{ij}^t$ and $\|e_{ij}^t\|_2$ represents the relative distances and the Euclidean distance between source node i and target node j at the current timestep t.

We also included additional features tailored to the metal forming case study, namely global feature vector $\boldsymbol{u}$ and contact feature vector $v_{ncont}^t$ in Table 1. In global feature vector, $\Delta t$ denotes the duration of each timestep interval in the simulation, stroke represents how much the punch moves vertically to deform the workpiece. Contact feature vector represents the contact interaction of rigid body forming tools and the deformable sheet metal object. The contact feature vector is a node-level feature vector that contains the inverse of node-to-surface distance $\frac{1}{d_n^t}$ and contact direction vector $\boldsymbol{n}^t$. The method used to compute contact feature vectors is described in Section 4.2. The model is trained to predict the node displacement vector $\Delta x_n^{t+1}$ from current timestep to the next timestep in autoregressive manner.

## 4.2 Contact modelling

As mentioned in Section 2.2, in the material forming, the tools (rigid body objects) are of comparable size to the blank (deformable object). Additionally, the rigid body objects are also in complex 3D shapes that require more detailed and accurate contact modelling method. To model the complex and large-scale rigid-deformable contact interactions, a node-to-surface contact modelling method is proposed inspired by the contact searching method in FE simulations [38]. The node-to-surface contact modelling method is shown in Fig. 4. At each timestep, the node-to-surface distance $d_n^t$ is computed for each deformable object nodes, defined as the shortest Euclidean distance to the surface of the rigid body. This distance value is then inverted as $\frac{1}{d_n^t}$, providing a simple yet informative measure of contact proximity. Additionally, the normal vector $\boldsymbol{n}^t$ of the closest rigid body object element is computed to represent the contact node features $\boldsymbol{v}_{\text{ncont}}^t$. The "closest rigid body object element" refers to the mesh element on the rigid body surface that is closest to the deformable body node in Euclidean distance. This normal vector provides directional information of the contact.

This method assigns all contact-related information directly to the nodes of the deformable object, ensuring no loss of critical interaction details. Since the features are constructed from relative positional information, the contact representation is naturally permutation-invariant. Furthermore, this method maps all contact related information on deformable object nodes. Finally, Appendix A compares the node-to-surface distance calculated using the proposed contact modelling method and those obtained from FE simulation results. The results demonstrated the agreement between the proposed contact modelling method and FE simulation results.

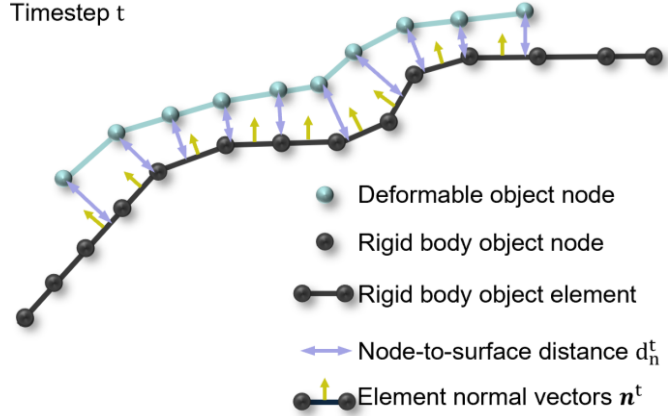

Fig. 4 Illustration of the proposed node-to-surface contact modelling method (view from 2D cross-section).

## 4.3 Graph coarsening method

As mentioned in Section 1, sheet formed components typically exhibit complex and large-scale 3D geometries. To represent these accurately, the corresponding graphs require large numbers of nodes and edges. Performing message passing on such large graphs is slow and computationally expensive. To address this, we adopt a hierarchical downsample/upsample mechanism in the model architecture, as illustrated in Fig. 5 (a) and detailly explained in Fig. 6 (b). During this process, information is gradually transferred to the coarser graph levels via inter-level edges. Message passing is then performed on the coarsest graph level. The updated node features are subsequently unsampled back to the finest graph level through the same set of inter-level edges. This mechanism reduces the number of message passing layers required to transfer information across long spatial distances. As a result, both memory consumption and computational cost are significantly reduced.

To enable the downsample/upsample mechanism, graph coarsening was applied to construct a hierarchy of coarse graphs from the original fine graph. Different graph coarsening methods were explored and compared to identify the most suitable one for the material forming case studies. Details of the methods and their comparative results are provided in Appendix B. We adopted a manual graph coarsening method. Blank meshes with varying densities were generated using commercial meshing software by adjusting the mesh element size. Each mesh was then used to define a corresponding graph. Lower mesh density corresponded to a coarser graph. This method ensured that the coarsened graphs accurately represented the overall shape of the geometry, which were particularly important for material forming case studies with complex boundary conditions and geometric features. This study focused on how different rigid tool geometries affect the deformation behaviour of a target deformable object, referred to as the blank. Each case study used a uniform blank shape and a consistent graph hierarchy. In the dome-shaped case study, three coarse graphs were constructed to gradually reduce the node count from 1,200 to 30. In the bulkhead-shaped case study, four coarse graphs were constructed to gradually reduce the node count from 6,700 to 40.

To support hierarchical message passing, we constructed inter-level edges between adjacent graph levels. These inter-level edges were generated using a KD-Tree-based inter-level mapping method. This method applied the KD-Tree algorithm to identify the three nearest coarse nodes for each fine node. Inter-level edges were constructed to link each fine node to its nearest coarse nodes. This formed inter-level edge sets between two adjacent graph levels. The inter-level mapping ensured full connectivity between the fine graph and all coarser levels, preserving spatial relationships without loss of geometric information. The mapping results for the dome-shaped case study and bulkhead-shaped case study are shown in Fig. 5 (b). Note that this section focuses on the data preparation. The use of inter-level edges for message passing will be detailed in Section 5.3.

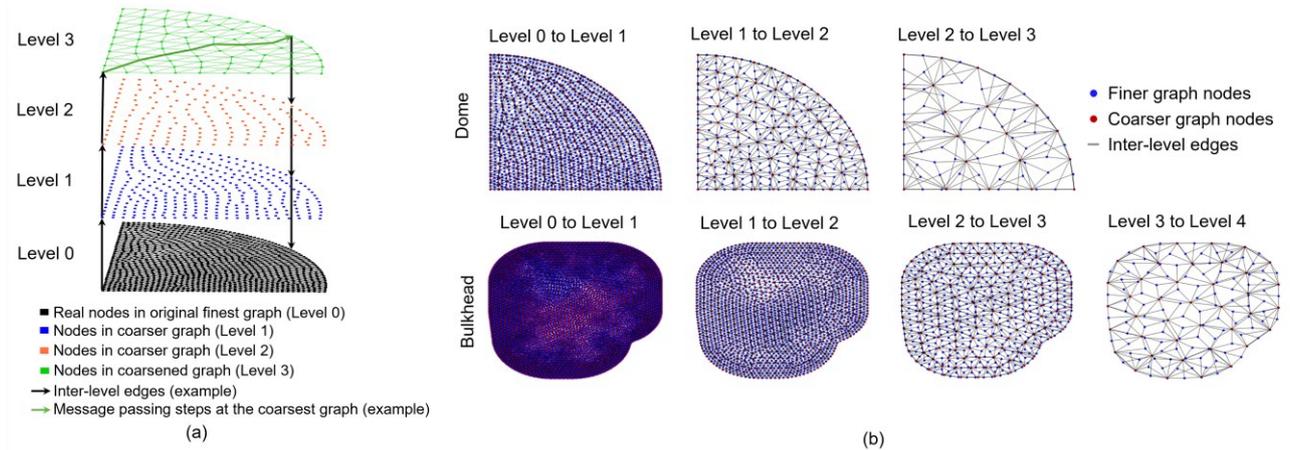

Fig. 5 (a) Graph downsampling/upsampling mechanism, illustrated using the dome-shaped case study; (b) inter-level mapping results for the two case studies.

## 5. RUGNN model development

### 5.1 Overall RUGNN framework

The RUGNN architecture is illustrated in Fig. 6. It follows an encoder–processor–decoder design. The encoder maps input graph feature vectors to latent space feature vectors. The processor adopts a downsample/upsample mechanism and three Recurrent-based GNN Processor Blocks (RGPBs). Finally, the decoder generates the displacement vector field at each timestep. Detailed descriptions of the encoder, processor, and decoder are provided in the following subsections.

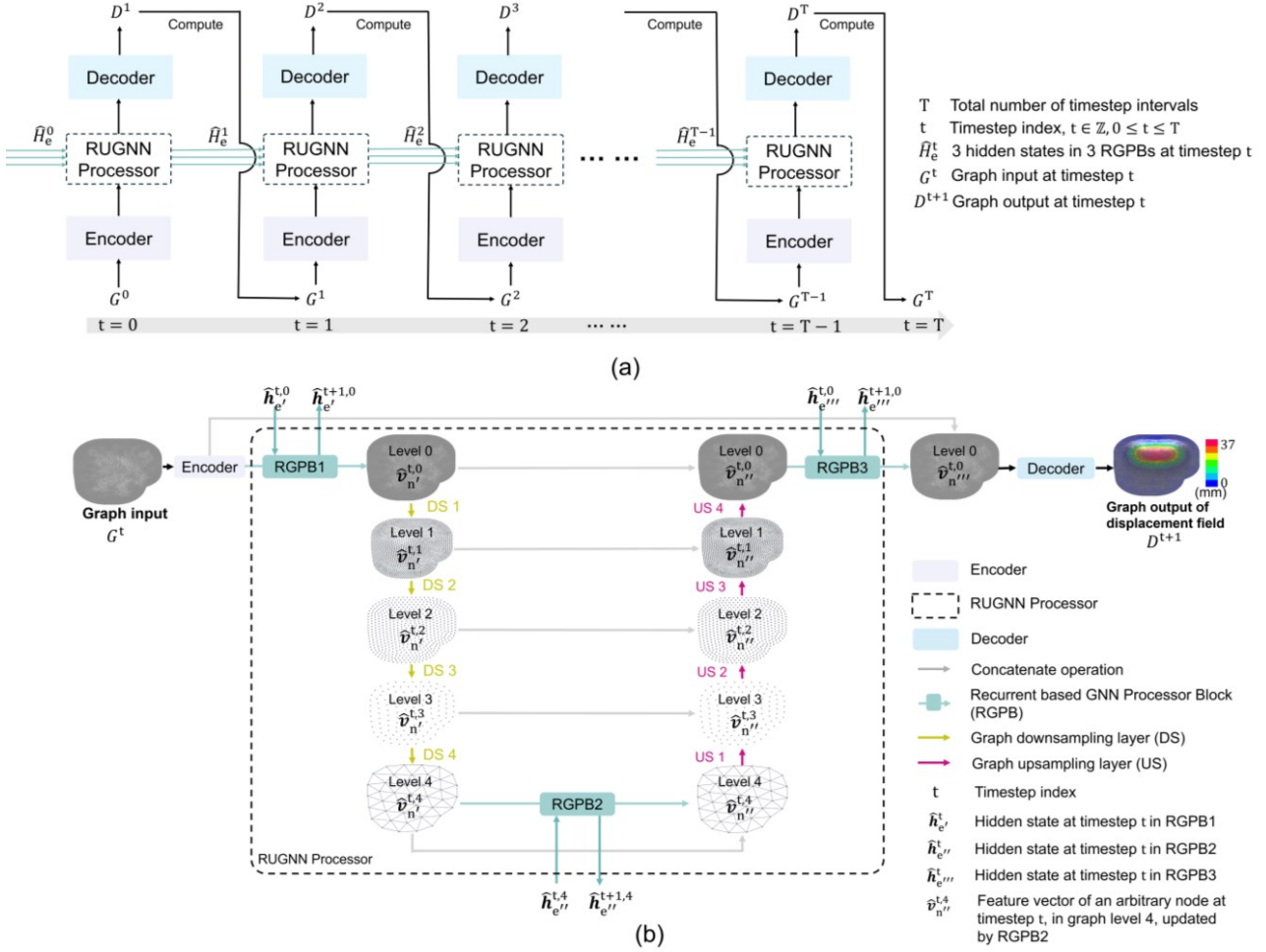

Fig. 6 (a) RUGNN for autoregressive prediction across timesteps; (b) The RUGNN framework at each timestep.

Fig. 6 (a) illustrates the RUGNN model used to autoregressively predict the displacement field over multiple simulation timesteps. The stamping simulations are discretised into T timestep intervals, resulting in T + 1 timesteps from the undeformed initial shape to the fully formed final shape. At every timestep $t \in \mathbb{Z}, 0 \leq t \leq T - 1$, the predicted displacement field $D^{t+1}$ is used to compute and update the graph input for the next timestep $t + 1$. This process continues until the fully formed final shape at timestep $t = T$ is obtained.

The detailed framework of RUGNN at a single timestep is shown in Fig. 6 (b). In this section, we introduce a unified notation convention to describe the feature vectors used in the RUGNN framework. The subscript indicates how a vector has been updated by the RGPBs. Specifically, $(*)_{*'}$, $(*)_{*''}$, and $(*)_{*'''}$ indicate the vectors have been updated by RGPB1, RGPB2, and RGPB3, respectively. The superscript indicates the graph level in the hierarchical graph structure during downsample or upsample. Specifically, $(*)^{t,0}$, $(*)^{t,1}$, $(*)^{t,2}$, $(*)^{t,3}$, and $(*)^{t,4}$ refer to variables at timestep on graph levels 0 to 4, respectively. For hidden state vectors, the subscript $(*)_{*'}$, $(*)_{*''}$, and $(*)_{*'''}$ indicate the hidden state belongs to RGPB1, RGPB2, RGPB3, respectively.

## 5.2 Encoder

In RUGNN, the encoder takes graph input ($G^t$) and converts them into a latent space as shown in Fig. 7 (a). In the encoder, the original node feature vector ($v_n^t$), contact feature vector ($v_{ncont}^t$) of an arbitrary node, edge feature vector ($e_e^t$) of an arbitrary edge, and global feature vector ($u$) are converted to latent space feature vectors, denoted as $\hat{v}_n^{t,0}$, $\hat{v}_{ncont}^{t,0}$, $\hat{e}_e^{t,0}$, and $\hat{u}$. These encoder Multilayer Perceptron (MLPs) consist of three linear layers. The first two layers are followed by a LeakyReLU activation function with a negative slope of 0.01, and the final layer is followed by a Layer Normalisation operation.

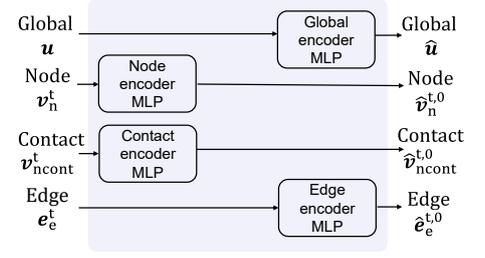

Fig. 7 Encoder design of RUGNN.

## 5.3 Processor

As shown in Fig. 6, the fundamental GNN architecture used in this study is the Message Passing Neural Network (MPNN)[11, 13, 39]. MPNN is a specialised GNN framework designed to update both node- and edge-level features though message passing. This architecture enables detailed information aggregation and precise node feature updates, improving prediction accuracy across various mechanics-related domains [15]. The following subsection provides a detailed discussion of the RUGNN processor, including its downsample/upsample mechanism, and RGPBs.

### 5.3.1 Downsample and upsample mechanism

As discussed in Section 4.3, graph-based downsample/upsample mechanism is adopted to enhance message passing efficiency. The downsample/upsample mechanism is illustrated in Fig. 6 (b) and Fig. 5. In this mechanism, node features at the target graph level in the graph hierarchy are updated via message passing from the source level, using the inter-level edges.

All downsample operations follow the same operation methodology, where the nodes in the finer graph level serve as source nodes and the nodes in coarser graph level serve as target nodes. We use Downsample layer 1 (DS1 in Fig. 6 (b)) as an example, its mathematical formulation is given by:

$$\hat{v}_{n_j'}^{t,1} = \sigma \left( \sum_{i \in \mathcal{N}(j)} \hat{v}_{n_i'}^{t,0} \cdot w_{e_{ij}} + b \right) \tag{1}$$

Here, $\hat{v}_{n_j'}^{t,1}$ denotes the feature vector of target node j in the coarser graph level (level 1), and $\hat{v}_{n_i'}^{t,0}$ is the feature vector of source node i in the finer graph level (level 0). $i \in \mathcal{N}(j)$ represents all neighbouring source nodes that are connected to target node j via inter-level edges. Each inter-level edge is associated with a learnable weight vector $w_{e_{ij}}$. During the downsample operation, the source node feature vectors are first transformed by their corresponding inter-level edge weight vectors. The weighted features are then aggregated at the target node, followed by the addition of a bias term $b$. Finally, a LeakyReLU activation function $\sigma$ is applied to obtain the updated target node feature vector $\hat{v}_{n_j'}^{t,1}$.

Similarly, all upsampling operations follow the same mechanism. However, they perform message passing in the reverse direction to downsample operations. In each upsample operation, the source nodes are taken from the coarser graph level, and the target nodes belong to the finer graph level. We use Upsampling layer 1 (US1 in Fig. 6(b)) as an example, its mathematical formulation given by:

$$\hat{v}_{n_j''}^{t,3} = \sigma \left( \sum_{i \in \mathcal{N}(j)} concat(\hat{v}_{n_i'}^{t,4}, \hat{v}_{n_i''}^{t,4}) \cdot w_{e_{ij}} + b \right) \tag{2}$$

Here, $\hat{v}_{n_j''}^{t,3}$ denotes the feature vector of target node j in the finer graph level (level 3). $concat(\cdot)$ denotes feature concatenation. $\hat{v}_{n_i'}^{t,4}$ is the feature vector of source node i obtained from DS4, while $\hat{v}_{n_i''}^{t,4}$ is the feature vector of source node i in the coarser graph level (level 4). $i \in \mathcal{N}(j)$ represents all neighbouring source nodes i that are connected to target node j via inter-level edges. Each inter-level edge is associated with a learnable weight vector $w_{e_{ij}}$. During upsample operation, the source node feature vectors are first concatenated and then transformed by their corresponding inter-level edge weight vectors. The resulting features are then aggregated at the target node, followed by the addition of a bias term $b$. A LeakyReLU activation function $\sigma$ is applied to obtain the updated target node feature vector $\hat{v}_{n_j''}^{t,3}$.

### 5.3.2 Recurrent based GNN Processor Block (RGPB)

The case studies involve predicting displacement fields at multiple timesteps, modelling the temporal relationships between different timesteps is essential. Therefore, three Recurrent-based GNN Processor Blocks (RGPBs) are used in this study, as shown in Fig. 6 (b). RGPB 1 operates on the finest graph before any downsample operations. It is designed to process and update initial node and edge features in a local scale before downsample operations. RGPB 2 operates on the coarsest graph structure after all downsample operations, responsible for propagating, aggregating, and updating edge and node information over spatial long distance via multiple stacked message passing layers. RGPB 3 works on the finest graph structure after all upsample operations. It performs local message passing to refine the updated node features before feeding them into the decoder. Despite their different roles, all RGPBs share the same internal structure. Fig. 8 provides an example of RGPB3.

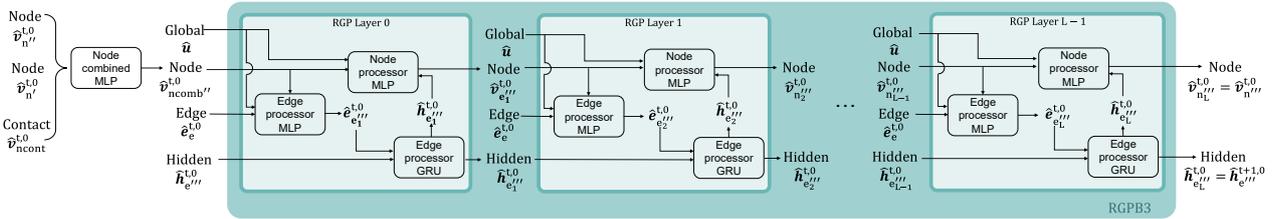

Fig. 8 Recurrent-based GNN Processor Block, illustrated using RGPB3 as example.

Inspired by a previous study [23], each RGPB consists of L number of Recurrent-based GNN Processor (RGP) layers. Each RGP layer contains separate set of network parameters and is applied in sequence to the output of previous RGP layer. The inputs to the first RGP layer (Layer 0) include the node feature vectors $\hat{v}_{n''}^{t,0}$ and $\hat{v}_{n'}^{t,0}$, contact feature vector $\hat{v}_{ncont}^{t,0}$, edge feature vector $\hat{e}_e^{t,0}$, and global feature vector $\hat{u}$. The node feature vectors and contact feature vectors are combined using a node combined MLP to produce the combined node feature vector $\hat{v}_{ncomb''}^{t,0}$, and is directly input to the first RGP layer. The output for RGPB3 is the updated hidden state vector $\hat{h}_{e'''}^{t+1,0}$ and the updated node feature vector $\hat{v}_{n'''}^{t,0}$. In Fig. 8, the subscript $(*)_{*_L'''}$ denotes the vector is updated by RGP layer (L-1) in RGPB 3. At the first timestep of the simulation, the initial hidden edge vector $\hat{h}_{e'''}^{t,0}$ is zero initialised. In the later timesteps the hidden edge vector is updated based on the output of RGPB in the previous timestep.

Each RGP layer follows a three-step computation, updating the edge feature vectors, hidden edge feature vectors, and node feature vectors. All RGP layers have the same structure, therefore the RGP layer (L-1) in Fig. 8 is used as an example. The mathematical formulations for RGP layer (L-1) in RGPB3 are summarised by Equation (3-5).

$$\hat{e}^{t,0}_{e_{ij_L}'''} = f_e\left(concat(\hat{v}^{t,0}_{n_{i_{L-1}}'''}, \hat{v}^{t,0}_{n_{j_{L-1}}'''}, \hat{e}^{t,0}_{e_{ij}}, \hat{u})\right) \quad (3)$$

Where, $n_i$ represents the source node i and $n_j$ represents the target node j, $e_{ij}$ represents the edge connecting source node i and target node j. The edge processor MLP ($f_e$) computes the updated edge feature vector $\hat{e}^{t,0}_{e_{ij_L}'''}$ for the edge $e_{ij}$. The input is the concatenation of the feature vector $\hat{v}^{t,0}_{n_{i_{L-1}}'''}$ of source node i, the feature vector $\hat{v}^{t,0}_{n_{j_{L-1}}'''}$ of target node j, the original edge feature vector $\hat{e}^{t,0}_{e_{ij}}$, and global feature vector $\hat{u}$. The node feature vectors with subscript $(*)_{*_{L-1}'''}$ denotes the feature vectors are updated based on the previous RGP layer (L-2). It is worth noting that, for RGPB2 operating on the coarsest graph level, edge feature vectors are calculated using the same approach described in Section 4.1.

$$\hat{h}^{t,0}_{e_{ij_L}'''} = f_{gru_e}\left(\hat{e}^{t,0}_{e_{ij_L}'''}, \hat{h}^{t,0}_{e_{ij_{L-1}}'''}\right) \quad (4)$$

In Equation (4), the edge processor GRU ($f_{gru_e}$) computes the updated hidden edge vector $\hat{h}^{t,0}_{e_{ij_L}'''}$ of each edge $e_{ij}$. The inputs include the updated edge feature vector $\hat{e}^{t,0}_{e_{ij_L}'''}$ from Equation (3), and a hidden edge vector $\hat{h}^{t,0}_{e_{ij_{L-1}}'''}$ updated by the previous RGP layer (L-2). The updated hidden edge vector of each RGP layer serves as the hidden edge vector input to the next RGP layer within the same timestep. After all L RGP layers are applied, the final hidden edge vector $\hat{h}^{t,0}_{e_L'''}$ of an arbitrary edge in the graph is retained and passed forward to the next timestep. Therefore, in Fig. 8, the output hidden edge vector $\hat{h}^{t+1,0}_{e'''}$ corresponds to the final hidden edge vector $\hat{h}^{t,0}_{e_L'''}$ of the last RGP layer (L-1) in RGPB3.

$$\hat{v}^{t,0}_{n_{j_L}'''} = f_v\left(concat(\sum_{e_{ij}\in\mathcal{N}_j} \hat{h}^{t,0}_{e_{ij_L}'''}, \hat{v}^{t,0}_{n_{j_{L-1}}'''}, \hat{u})\right) \quad (5)$$

In Equation (5), the node processor MLP $f_v$ computes the updated node feature $\hat{v}^{t,0}_{n_{j_L}'''}$ of target node j. This computation consists of an aggregation operation followed by a node update operation. In the aggregation operation, $e_{ij} \in \mathcal{N}_j$ represents the edges $e_{ij}$ that are connected to target node j. The hidden edge vectors $\hat{h}^{t,0}_{e_{ij_L}'''}$ of edges $e_{ij}$ are aggregated via element-wise summation. In the node update operation, the aggregated hidden edge vector ($\sum_{e_{ij}\in\mathcal{N}_j} \hat{h}^{t,0}_{e_{ij_L}'''}$) is concatenated with the self-loop target node feature vector $\hat{v}^{t,0}_{n_{j_{L-1}}'''}$ and a global feature vector $\hat{u}$. The resulting concatenated vector is then passed through the node processor MLP $f_v$ to update the target node feature vector $\hat{v}^{t,0}_{n_{j_L}'''}$. The updated node feature vector of each RGP layer serves as the node feature vector input to the next RGP layer within the same timestep. After all L RGP layers are applied, the final updated node feature vector $\hat{v}^{t,0}_{n_L'''}$ of an arbitrary node in the graph is retained and passed forward to the decoder. Therefore, in Fig. 8, the output node feature vector $\hat{v}^{t,0}_{n'''}$ corresponds to the final hidden state $\hat{v}^{t,0}_{n_L'''}$ of the last RGP layer in RGPB3.

Overall, the inclusion of a GRU layer in each RGPB allows the model to incorporate temporal information from previous timesteps into the current timestep. Meanwhile, this temporal information is spatially aggregated and updated through message passing, enabling the model to capture spatial-temporal dependencies.

## 5.4 Decoder

The decoder MLP is applied to the final updated node features after RGPB3, as shown in Fig. 6 (b). The detailed design of decoder is illustrated in Fig. 9. The input of decoder is the updated node feature vector from the RGPB3 $\hat{v}_{n'''}^{t,0}$, the original node feature vector from encoder $\hat{v}_n^{t,0}$, and the original contact feature vector $\hat{v}_{ncont}^{t,0}$. The output of decoder MLP is the node displacement vector $\Delta x_n^{t+1}$ at each timestep t. The decoder MLP consists of three linear layers. The first two layers are followed by a LeakyReLU activation function with a negative slope of 0.01, and the final layer is followed by a Layer Normalisation operation.

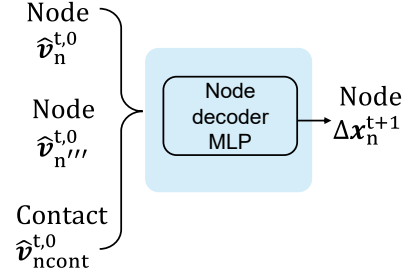

Fig. 9 Decoder design of RUGNN.

## 6. Training strategies

The models used in the dome-shaped case study were trained on a workstation featuring two Nvidia Quadro 5000 GPUs, with a total of 32GB of GPU memory. The models used in the bulkhead-shaped case study were trained on Google Colab, using an Nvidia Tesla A100 GPU with 40GB of GPU memory. All models were trained using the Adam optimiser, with a scheduled learning rate that decayed exponentially from $\exp(-4)$ to $\exp(-6)$. The training loss and validation loss were formulated using the Mean Squared Error (MSE), as defined in Equation (6):

$$\mathcal{L}_{MSE} = \frac{1}{3MN}\sum_{m=0}^{M-1}\sum_{t=0}^{T-1}\sum_{n=0}^{N-1}\sum_{d\in\{x,y,z\}}\left(p_{n,d}^t - g_{n,d}^t\right)^2 \tag{6}$$

Where $\mathcal{L}_{MSE}$ represents the MSE loss accumulated over all timestep intervals, M represents the total number of samples in dataset, T represents the total number of timestep intervals, 10 in both case studies, t represents the timestep index. N represents the total number of nodes in the graph at each timestep. $d \in \{x, y, z\}$ represents the x, y, and z direction components in the cartesian coordinate system. $p_{n,d}^t$ and $g_{n,d}^t$ represent the predicted and ground truth displacement value for node n, in direction d, at timestep t. $\frac{1}{3MN}$ indicates that $\mathcal{L}_{MSE}$ is averaged over all 3 directions in cartesian coordinate, all samples, and all graph nodes.

Both teacher forcing training strategy and the autoregressive training strategy were adopted. The teacher forcing training takes ground truth graph input feature at current timestep and predicts the displacement vector fields at the corresponding timestep. In contrast, the autoregressive training strategy uses the predicted displacement vector field and the updated graph input features from the previous timestep to predict the displacement vector field at the current timestep. This autoregressive process as illustrated in Fig. 6 (a). After each prediction, the output is fed back to compute the graph input features for the next timestep. This iterative process continues until the final timestep, resulting in a sequence of predicted displacement vector fields $D^1, D^2, \ldots, D^T$.

## 7. Model performance evaluation

In this section, the performance of our proposed model is compared with several GNN baselines. Specifically, we consider two distinct case studies: the dome-shaped cold stamping case study, and the bulkhead-shaped hot stamping case study. Section 7.1 describes the comparison results for dome-shaped cold stamping case study, Section 7.2 describes the comparison results for the bulkhead-shaped hot stamping case study, and Section 7.3 compares the RUGNN performance under different hyperparameters, model training strategies, and input feature configurations.

### 7.1 Baseline comparison on the dome-shaped case study

The performance of the proposed RUGNN model was compared against three baseline models. (1) VanillaGNN: an MPNN without RGPBs and downsample/upsample mechanism. (2) RGNN: an MPNN with a RGPB on the finest graph level but without downsample/upsample mechanism. (3) UGNN: an MPNN with downsample/upsample mechanism, but without RGPBs. Detailed specifications for the baseline models and RUGNN are provided in Appendix C. The performance of the models was evaluated using various evaluation metrics, as illustrated in Fig. 10.

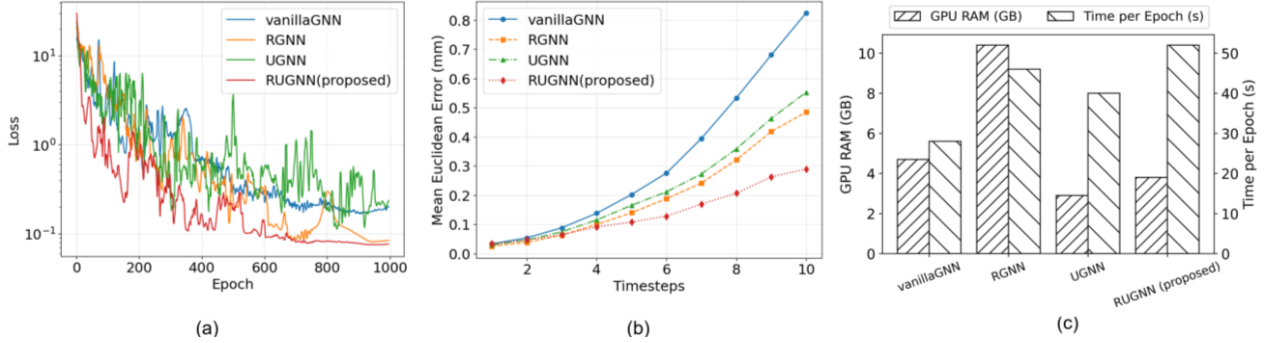

Fig. 10 (a) Autoregressive validation loss curves; (b) The Mean positional Euclidean Error (MEE) between the ground truth graph and the predicted graph, evaluated across all test samples in the dome-shaped case study; (c) Model efficiency evaluation.

Fig. 10 (a) compares the roll-out validation loss curves of the RUGNN model and three baseline models in the dome-shaped case study. Fig. 10 (b) further compares the positional mean positional Euclidean errors from timestep 1 to the final timestep, calculated using Equation (7):

$$\text{MEE}^t = \frac{1}{MN}\sum_{m=0}^{M-1}\sum_{n=0}^{N-1}\sqrt{\sum_{d\in\{x,y,z\}}(x_{n,d}^t - x^{GT^t}_{n,d})^2} \quad (7)$$

where $\text{MEE}^t$ is the mean positional Euclidean error at timestep $t \in \mathbb{Z}, 1 \leq t \leq 10$, covering the predicted deformation from the first deformed shape at timestep 1 to the fully formed final shape at timestep 10. M represents the number of test samples in the test dataset, N represents the total number of nodes in the graph at each timestep. $d \in \{x, y, z\}$ represents the x, y, and z direction components in the cartesian coordinate system. $x_{n,d}^t$ represents the predicted position value for node n in direction d at timestep t, $x^{GT^t}_{n,d}$ represents the ground truth position value for node n, in direction d, at timestep t. Noted that the predicted position $x_{n,d}^t$ is computed based on the predicted displacement values $p_{n,d}^t$ in Equation (6). At timestep 1, it is obtained by adding the predicted displacement at timestep 0 to the initial undeformed node position. At timestep 2, it is computed by adding the predicted displacement at timestep 1 to the previously computed position at timestep 1. This iterative process continues until the final timestep.

From Fig. 10 (b), the vanillaGNN model exhibits the highest accumulated autoregressive mean positional Euclidean error. RGNN shows significantly lower roll-out accumulated error than vanillaGNN due to the incorporation of recurrent blocks in the processor layer. These recurrent blocks integrate historical temporal data from previous timesteps. When combined with the MPNN architecture, they enable the model to update spatial and temporal features, making it more suitable for autoregressive predictions across multiple incremental timesteps. Additionally, UGNN with downsample/upsample mechanism also outperforms vanillaGNN. This result highlights the importance of sufficient message passing across long spatial distances. The proposed RUGNN model with downsample/upsample mechanism and recurrent blocks achieves the lowest positional MEE across all timesteps and reduces error accumulation compared to all baseline models.

Fig. 10 (c) evaluates the computational efficiency of the baseline models and the RUGNN model. Due to the downsample/upsample mechanism, RUGNN significantly reduces GPU consumption compared to vanillaGNN and RGNN. However, the introduction of 3 recurrent blocks in the processor layer slightly increases training time compared to baseline models. Given its high predictive accuracy, RUGNN achieves a favourable balance between computational efficiency and accuracy, making it a practical choice for autoregressive predictions in large-scale simulations.

The prediction results of baseline models and RUGNN are visualised on two test cases featuring the largest dome radius in Fig. 11. Positional Euclidean error is used to measure overall prediction accuracy between the ground truth position and predicted position in the final timestep.

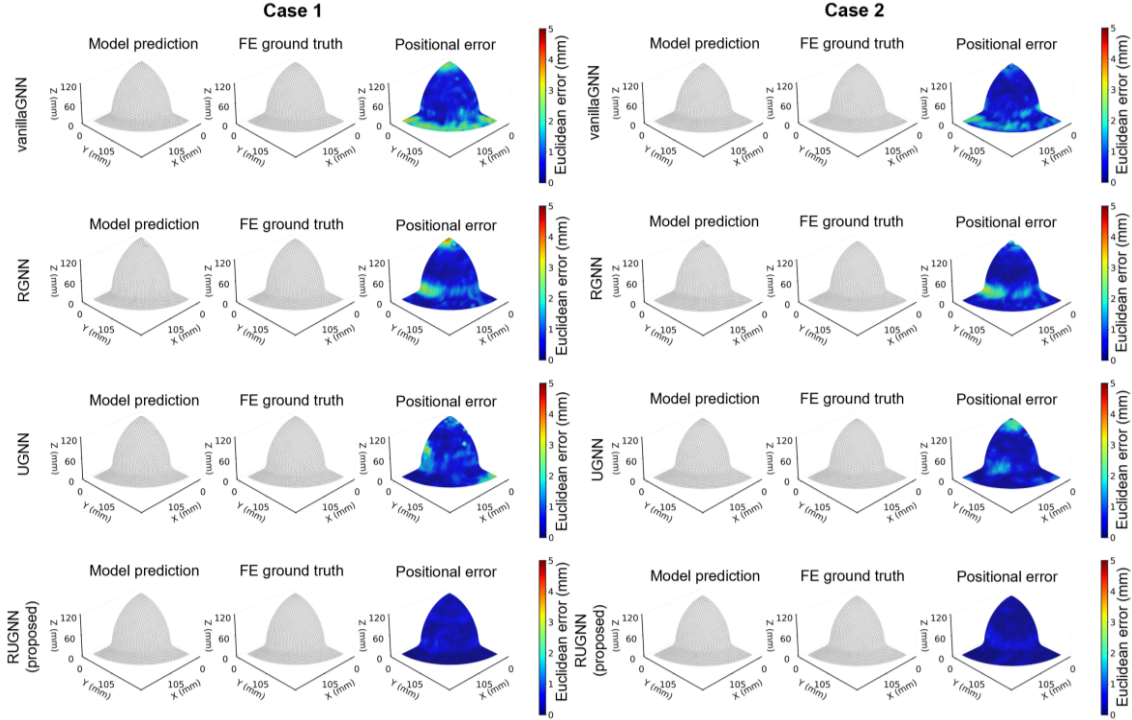

Fig. 11 Comparison of FE ground truth and model prediction for two test cases.

According to Fig. 11, both vanillaGNN and RGNN exhibit high localised errors in the top and bottom regions. These two models perform message passing on the finest graph level without downsampling/upsampling mechanism. This suggests that using the current setup of message passing layers is insufficient to propagate information over longer spatial distances. Increasing the number of passing layers may mitigate this problem; however, this method would significantly increase GPU memory consumption and training time. This makes such models impractical for large-scale graphs with long-range dependencies. In contrast, the UGNN and RUGNN models, which incorporate the downsampling/upsampling mechanism, demonstrate better performance in the top and bottom regions. In these models, message passing is primarily performed on the coarsest graph level, where a moderate number of layers is sufficient to propagate information over longer spatial distances. This improves the model's ability to learn long-range dependencies induced by material stiffness, while maintaining reasonable computational costs.

Additionally, vanillaGNN and UGNN exhibit more scattered large Euclidean errors compared to RGNN and RUGNN. This may be attributed to the absence of recurrent blocks. As previously discussed, models without recurrence block(s) tend to have higher accumulated errors over time. In vanillaGNN and UGNN, small errors are more likely to be amplified during autoregressive rollout, resulting in more scattered high errors in the final timestep.

Finally, RUGNN, which integrates both RGPBs and the downsampling/upsampling mechanism, have the highest predictive accuracy. Its predictions closely match the FE simulation ground truth. The full sequence of predicted positions, from the first deformed shape to the fully formed final shape, is provided in Appendix D, demonstrating that RUGNN maintains high prediction accuracy throughout all timestep intervals.

## 7.2 Baseline comparison on the bulkhead-shaped case study

To evaluate the RUGNN model performance on more complex and larger-scale setups, we compared RUGNN with several baseline models using a bulkhead-shaped dataset. Each sample in the bulkhead-shaped dataset contains 6,700 nodes, which is significantly larger than the dome-shaped case study with 1,300 nodes. Additionally, the bulkhead-shaped case study is based on hot stamping simulations using elastic-viscoplastic material with temporal dependant material properties, which adds further non-linearity to the spatio-temporal displacement field prediction. This evaluation is crucial for assessing RUGNN's performance in handling large-scale, highly nonlinear deformation processes.

The RUGNN model was compared against two baseline models: (1) vanillaGNN: an MPNN without RGPBs or downsample/upsample mechanism. (2) UGNN: an MPNN with downsample/upsample mechanism, but without RGPBs. Detailed specifications for the baseline models and RUGNN are provided in Appendix E. All models were trained for 1000 epochs. The performance of the models was evaluated using various evaluation metrics, as illustrated in Fig. 12.

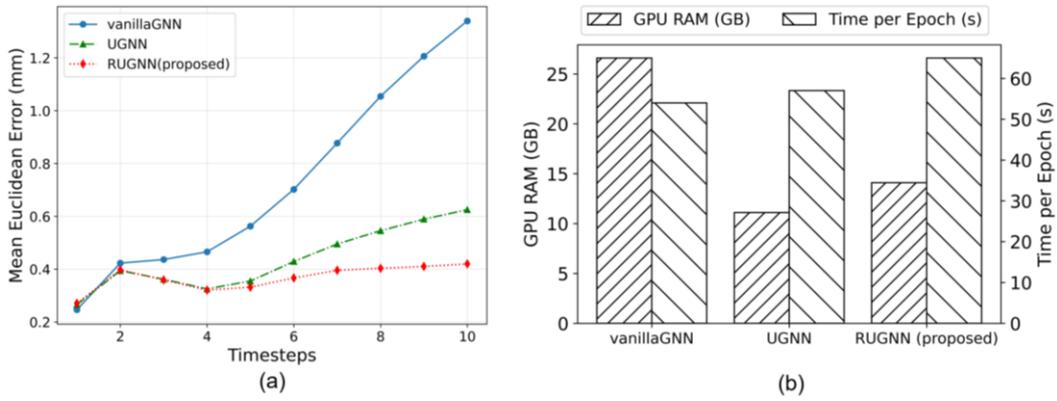

Fig. 12 (a) The Mean Euclidean Error (MEE) between the ground truth graph and the predicted graph, evaluated across all test samples in the bulkhead-shaped case study; (b) Model computation efficiency evaluation.

Fig. 12 (a) compares the mean positional Euclidean errors $MEE^t$ from timestep 1 to the final timestep, This evaluation covers evaluates the positional predictive accuracy from the first deformed shape to the fully formed final shape. According to Fig. 12 (a), vanillaGNN has the highest accumulated positional error. UGNN model significantly reduces the roll out accumulated error compared to vanillaGNN model. Compared with UGNN model, the RUGNN model further reduces accumulated rollout error at later timesteps. Fig. 12 (b) compares the training efficiency of the baseline models and RUGNN. vanillaGNN, which operates on the finest graph without downsample/upsample mechanism, has the highest GPU memory consumption. UGNN model have the lowest GPU memory consumption due to the adoption of downsample/upsample mechanism. RUGNN model with additional recurrent blocks have slightly larger GPU consumption and training time compared with UGNN model. However, given its significant accuracy improvement, RUGNN achieves a favourable balance between computational efficiency and prediction accuracy, making it well-suited for autoregressive predictions in large-scale simulations.

To further evaluate the model performance, Fig. 13 visualises prediction results of baseline models and RUGNN model for two representative test samples. Compared to UGNN, vanillaGNN shows large prediction errors concentrated in the lower region. This suggests that using current setup of message passing layers in vanilla GNN is insufficient for information to propagate over longer spatial distances, leading to localised prediction errors in the bottom region. Increasing the number of message-passing layers in vanillaGNN may help alleviate this issue, but it would also lead to significantly higher GPU memory consumption and training time, making the model impractical for large-scale graph applications. In contrast, the UGNN model incorporates a downsampling/upsampling mechanism, where message passing is primarily performed on the coarsest graph level. With a moderate number of message-passing layers at the coarsest level, UGNN can effectively propagate information over longer spatial distances, making it more suitable for large-scale graph processing.

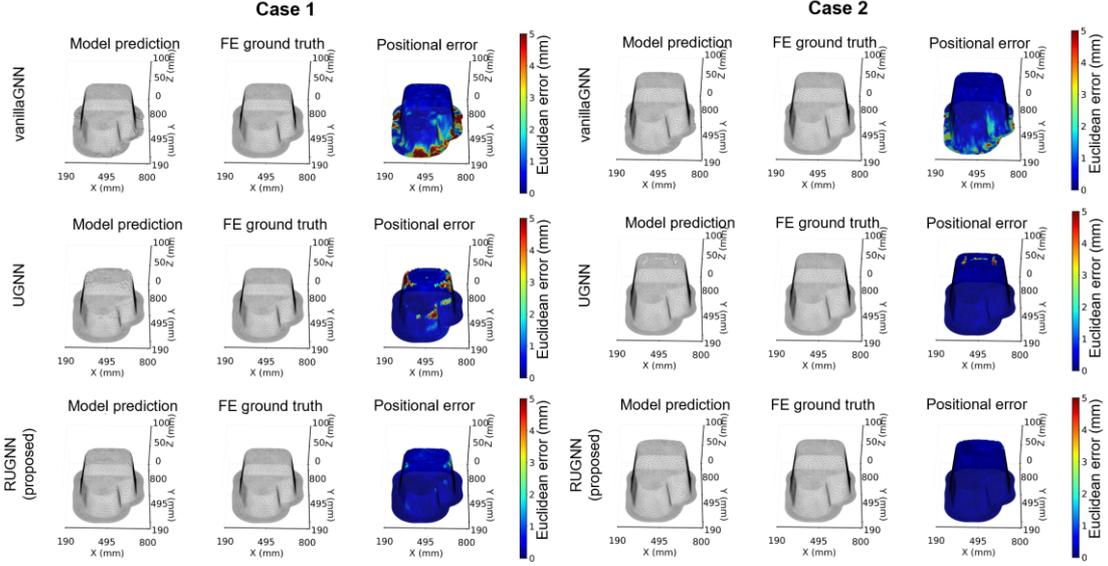

Fig. 13 Comparison of FE ground truth and network predictions for two bulkhead-shaped test cases.

Additionally, UGNN model have scattered large error in the top region compared to vanillaGNN and RUGNN model. There are two possible causes for this issue. First, vanillaGNN operates on the original finest graph level, and due to the limited number of message-passing layers, it tends to focus more on learning local features and capture fine-scale details. In contrast, UGNN primarily performs message passing on the coarsest graph level, which improves its ability to learn long-range spatial dependencies but reduces its capacity to accurately capture fine local details. Second, since UGNN does not incorporate recurrent blocks compared to RUGNN model, small errors are more likely to accumulate and amplify during autoregressive rollouts. This issue is particularly pronounced in the top region, where complex contact interactions with stamping tools introduce additional instability in predictions.

These findings highlight the complementary roles of the downsample/upsample mechanism and the recurrent block. While downsample/upsample mechanism enables efficient information propagation over longer spatial distance, the recurrent block helps mitigate error accumulation over time. Integrating both downsample/upsample mechanism and recurrent block, as in the RUGNN model, offers a robust solution for spatio–temporal prediction in large-scale graph-based simulations. The full sequence of predicted positions, from the first deformed shape to the fully formed final shape, is provided in Appendix F. These results demonstrate that RUGNN maintains high prediction accuracy across all timestep intervals in the bulkhead-shaped case study.

### 7.3 Model configuration analysis of RUGNN

This section presents a comprehensive analysis of different configurations on RUGNN model performance, including hyperparameter settings, training strategies, and input feature design. For the hyperparameter settings analysis, we explored variations in the number of message-passing layers, the number of downsample/upsample layers, node reduction ratios, and batch sizes to evaluate their impact on model performance. For the training strategy analysis, we evaluated teacher forcing and autoregressive training strategy on model performance. For the input feature design analysis, we investigated the effect of contact features and different initialisations of the hidden state. All evaluations are based on autoregressive roll out validation MSE loss, calculated using Equation (6). The results are summarised in both Fig. 14 and Table 2.

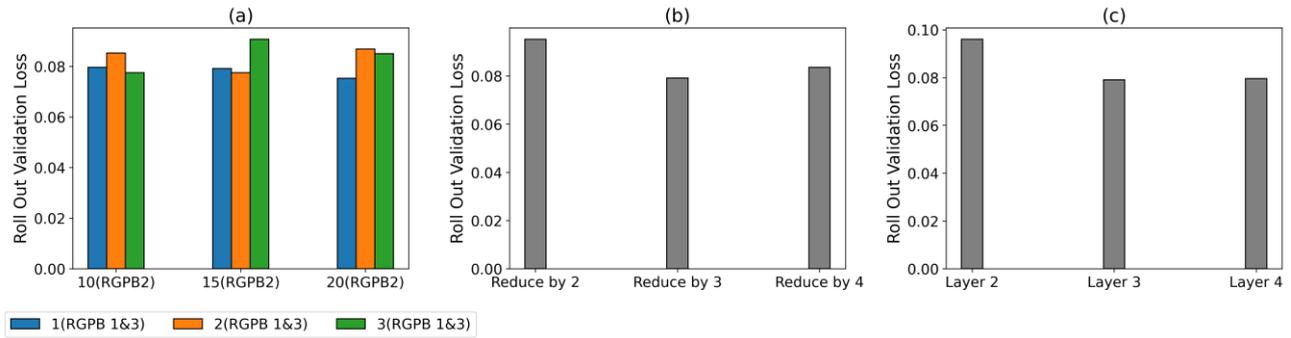

Fig. 14 Model performance under various factors (a) different message-passing layers; (b) different node reduction ratios; (c) different numbers of downsample/upsample layers.

Table 2 Impact of Boolean factors on autoregressive roll out validation loss.

| Batch size | | Training strategy | | Contact | | Initial hidden states | |
|---|---|---|---|---|---|---|---|
| 1 | **0.0791** | Teacher forcing | **0.0791** | With contact | **0.0791** | zero | **0.0752** |
| 2 | 0.1008 | Autoregressive | 0.0879 | Without contact | 0.0974 | Global feature | 0.0791 |

Regarding hyperparameter analysis, Fig. 14 (a) highlights the critical role of message-passing layers in the processor for model performance. The performance of RUGNN under various combinations of message passing layers is relatively robust. The message passing layers in the finest graph level (RGPB1 and RGPB3) primarily aggregates and propagate information locally between neighbouring nodes. The message passing layers in the coarsest graph level (RGPB2) primarily propagates information over longer spatial distances. Since RGPB1 and RGPB3 operates on the finest graph level, increasing the number of message passing layer in RGPB1 and RGPB3 would cause significant increase in GPU memory consumption during training. Considering the model efficiency, setting a message passing layer of 1 in RGPB1 and RGPB3 have the overall best model performance. Based on this configuration, the message passing layer is set to be 20 in RGPB2 to obtain the overall best model predictive performance.

Fig. 14 (b) and Fig. 14 (c) evaluate two critical aspects of the downsample /upsample mechanism. The node reduction ratio determines the proportion of coarser graph nodes to finer graph nodes between each two adjacent downsample/upsample graph levels. The number of downsample/upsample layers is directly related to the number of hierarchical graph levels used to represent the data. From Fig. 14 (b), setting the node reduction ratio to 3 offers a balance between preserving spatial information while maintaining computational efficiency. Based on this configuration, additional experiments were conducted to evaluate model performance under different downsample/upsample layers, as shown in Fig. 14 (c). For the dome-shaped case study, 3 downsample/upsample layers are rendered sufficient for message propagation. For different case studies, a node reduction ratio of 3 is recommended, the number of downsample and upsample layers can be adjusted based on the total number of mesh nodes in the graph. Furthermore, the first column in Table 2 examines the impact of batch size on model performance. It is discovered that using batch size 1 tend to have better result on autoregressive predictions.

Regarding training strategies, Table 2 shows that using the teacher forcing training strategy leads to lower autoregressive validation loss compared to autoregressive training strategy. The results related to input feature design are also summarised in Table 2. Training without contact information leads to significantly higher validation loss. This finding further supports the effectiveness of the contact representation method used in this study. Additionally, experiments show that using zero initialisation for the hidden state vectors yields slightly better performance than initialising with the global feature vector $\hat{\boldsymbol{u}}$.

## 8. Conclusions

This study proposed a novel Recurrent U-Net-based Graph Neural Network (RUGNN) framework for accurate prediction of material deformation in sheet material stamping across multiple timesteps. The key conclusions of this study are as follows:

- The RUGNN framework incorporates Gated Recurrent Units (GRUs) within its message-passing layers to effectively model temporal dynamics. It employs a graph-based U-Net architecture with a downsample/upsample mechanism, enabling efficient handling of long-range dependencies in large-scale, complex geometries.
- A new node-to-surface contact modelling method was proposed. This method is computationally efficient and robust in processing large-scale contact interactions.
- The RUGNN was validated on two sheet material forming case studies: aluminium cold stamping of dome-shaped geometries and aluminium hot stamping of bulkhead-shaped geometries. RUGNN achieved the highest prediction accuracy compared with baseline models.
- The RUGNN framework supports tool design exploration, allowing designers to assess manufacturability and develop complex 3D tool geometries for sheet metal forming.

Future research will focus on enhancing the framework's predictive capabilities to incorporate additional manufacturability factors, including thinning fields, von Mises stress, and other critical performance metrics. Additionally, we aim to further explore the model's capabilities over longer timesteps. Furthermore, we plan to develop an optimisation model that integrates the RUGNN framework with optimisation algorithms, enabling automated design iterations to identify optimal component geometries and process parameters. This advancement will streamline the design workflow, facilitating the development of lightweight, high-performance components.

## Author contributions: CRediT

Yingxue Zhao: Conceptualisation, Methodology, Data curation, Software, Visualisation, Writing-original manuscript. Qianyi Chen: Methodology, Writing– review & editing. Haoran Li: Methodology, Writing– review & editing. Haosu Zhou: Methodology, Writing– review & editing. Hamid Reza Attar: Data curation, Writing – review & editing. Tobias Pfaff: Supervision, writing – review & editing. Tailin Wu: Writing – review & editing. Nan Li: Conceptualisation, Funding acquisition, Investigation, Methodology, Project administration, Supervision, Writing – review & editing.

## Data availability

The data and codes will be made available upon reasonable request.

## Acknowledgement

The authors would like to thank ESI Group for their technical support with the PAM-STAMP software. Yingxue Zhao, Haoran Li and Haosu Zhou acknowledge the PhD Scholarships from Imperial College London.

# Appendix A Comparison of node-to-surface distance computed from proposed contact method and simulation results

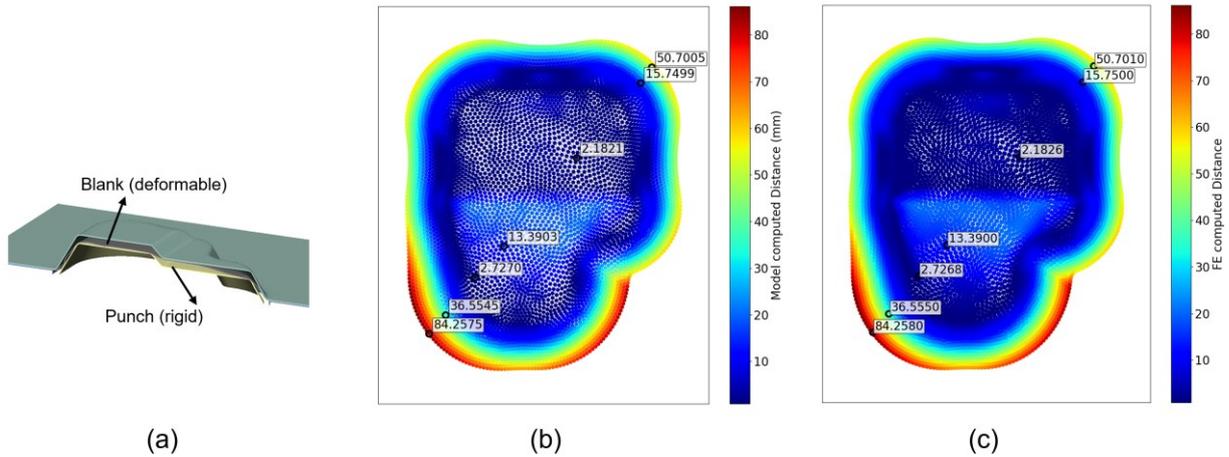

Fig.A.1 (a) illustration of the Blank and Punch position in an arbitrary timestep of an arbitrary sample; (b) Node-to-surface distance computed by proposed contact method; (c) Node-to-surface distance computed by FE simulations.

Appendix A presents a comparison between the node-to-surface distances computed using the proposed contact modelling method and those obtained from FE simulation results. The comparison is conducted on an arbitrary timestep of a randomly selected sample. To further assess the agreement between the proposed method and FE simulations, seven representative nodes were selected, and their corresponding node-to-surface distance values are shown in Fig. A.1(b) and Fig. A.1(c). The results show that the proposed method computes distance values closely aligned with the FE results, demonstrating its accuracy and effectiveness.

# Appendix B Comparison of different graph coarsening methods

To find the most suitable graph coarsening approach, four different graph coarsening approaches were compared and evaluated. The first three are automated graph coarsening approaches that do not rely on manual generations. These include edge-based clustering, node-based clustering, and Bi-stride graph coarsening methods. The fourth approach relies on manual generation of coarser meshes using commercial meshing software.

We experimented with an edge-based clustering method named Graph Partition Algorithm (gPartition) [34]. gPartition begins by identifying an initial edge within the graph of the fine mesh. The two nodes connected by this edge are merged into a single node. Subsequently, both the edge and its associated nodes are removed from the graph. This process is repeated iteratively, selecting and processing edges until all edges in the original graph have been evaluated. In the resulting coarsened graph, edges are reconstructed based on the connectivity of the merged nodes. Specifically, if two nodes in the coarse graph share neighbouring connections in the original fine graph, an edge is established between them in the coarsened graph.

We experimented with a node-based clustering method named G-pooling [31]. G-pooling starts by selecting an initial node index in the fine graph. This node, along with all its neighbouring nodes, forms the first cluster. Once the cluster is created, the process continues by selecting the next unprocessed node. For each subsequent node, a new cluster is formed by including the node and all its neighbours. This iterative procedure is repeated until all nodes in the fine graph have been assigned to a cluster. When constructing the coarse graph, edges are established between clusters based on the connectivity in the fine graph. Specifically, if any node in one cluster is connected to any node in another cluster in the fine graph, an edge is created between the corresponding nodes in the coarse graph.

We experimented with a Bi-stride graph coarsening method [40]. The Bi-stride graph coarsening method starts by selecting a seed node within the graph of the fine mesh. Using Breadth-First Search (BFS), the geodesic distances from the seed node to all other nodes are computed, assigning each node a depth level. Nodes are then categorised into odd layers and even layers based on odd and even depth assigned at each node. Typically, the nodes in the even layers are retained to form the coarsened graph. In the coarsened graph, edges are reconstructed based on the connectivity in the original fine graph. Specifically, if any two nodes in the coarse graph share a common neighbour in the original fine graph, a direct edge is created between them. Edges between even-layer nodes that already exist in the fine graph are preserved. This ensures that the coarse graph maintains sufficient connectivity and accurately represents the topology of the original fine graph [40].

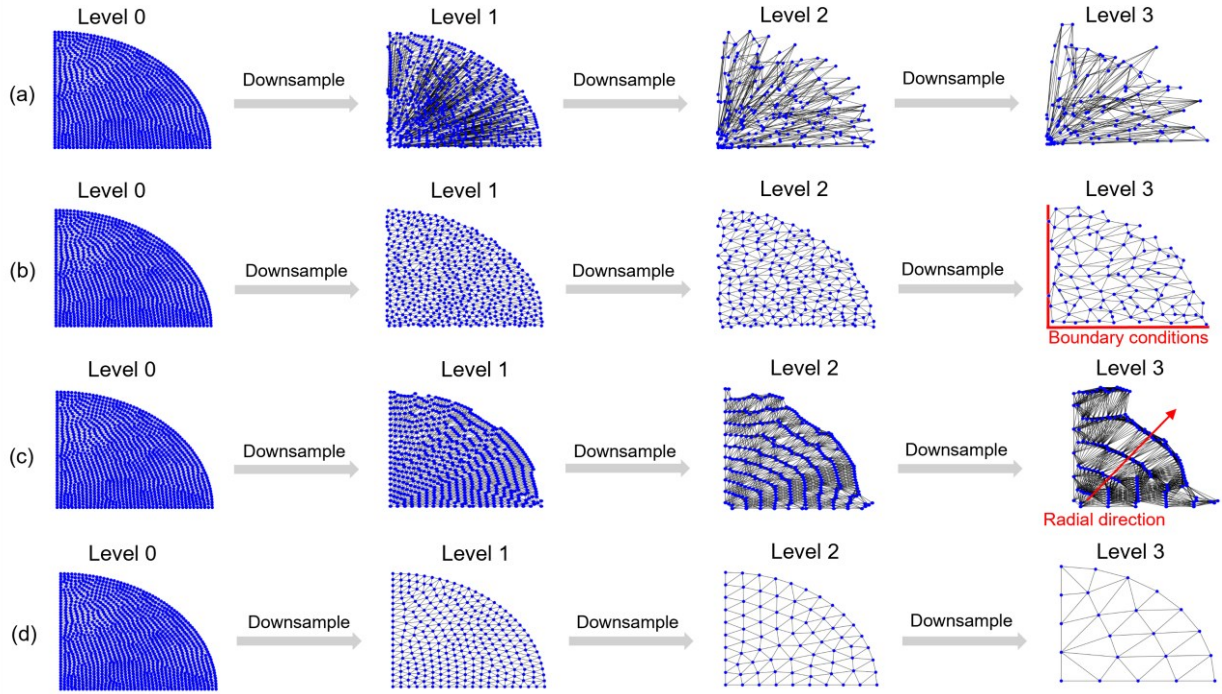

Fig.B.1 This figure compares four types of downsample strategies on the workpiece (blank) mesh in dome-shaped case study (a) edge-based clustering downsample strategy (gPartition); (b) node-based clustering downsample strategy (G-pool); (c) BFS-based downsample strategy (Bi-stride); (d) manual-based downsample using HyperMesh software.

Fig.B.1 compares four graph coarsening methodologies across multiple layers. The edge-based clustering method merges nodes connected by edges to reduce the graph complexity. This approach preserves local connectivity and captures fine-grained features in the original graph level (Level 0). However, as the graph is further coarsened (Level 2 and 3), the resulting structure becomes less regular, potentially impacting the global geometric representation.

Node-based clustering simplifies the graph by merging nodes and their neighbours, producing a more uniform graph structure. This method achieves high geometric regularity, as seen in Level 2 and 3. However, boundary node preservation after graph coarsening is suboptimal, as seen in the final coarsened graph.

BFS-based Bi-stride balances topology preservation and geometric regularity, ensuring smooth transitions and robust connectivity across graph coarsening layers. However, as the number of graph coarsening layers increases, the outer node layer fails to accurately represent the topology's shape, and the mesh edges appear excessively elongated in the radial direction, potentially losing spatial details along radial direction.

Overall, the manual-based graph coarsening method provides the most regular structure that preserves the overall shape of the geometry, particularly in deeper graph coarsening layers.

# Appendix C Model inputs and hyperparameters in dome-shaped case study

Table C.1 Detailed comparison of model inputs and hyperparameters in dome-shaped case study

| Model | Hyperparameters | | | Downsample/upsample |
|---|---|---|---|---|
| | **Processor block 1** | **Processor block 2** | **Processor block 3** | **Graph hierarchy** |
| **vanillaGNN** | Message passing layer = 10<br>Edge processor MLP (128, 128)<br>Node processor MLP (128, 128) | NA | NA | NA |
| **RGNN** | (RGPB)<br>Message passing layer = 10<br>Edge processor MLP (128, 128)<br>Edge processor GRU (128, 128)<br>Node processor MLP (128, 128) | NA | NA | NA |
| **UGNN** | Message passing layer = 2<br>Edge processor MLP (32, 32)<br>Node processor MLP (32, 32) | Message passing layer = 10<br>Edge processor MLP (128, 128)<br>Node processor MLP (128, 128) | Message passing layer = 2<br>Edge processor MLP (32, 32)<br>Node processor MLP (32, 32) | 4 levels<br>(32, 32, 64, 128) |
| **RUGNN (proposed)** | (RGPB 1)<br>Message passing layer = 2<br>Edge processor MLP (32, 32)<br>Edge processor GRU (32, 32)<br>Node processor MLP (32, 32) | (RGPB 2)<br>Message passing layer = 10<br>Edge processor MLP (128, 128)<br>Edge processor GRU (128, 128)<br>Node processor MLP (128, 128) | (RGPB 3)<br>Message passing layer = 2<br>Edge processor MLP (32, 32)<br>Edge processor GRU (32, 32)<br>Node processor MLP (32, 32) | 4 levels<br>(32, 32, 64, 128) |

Table C.1 summarises the input features, key hyperparameters, and architectural details of the baseline models and the proposed RUGNN. For vanillaGNN and RGNN, all processor units take the latent space representation of 128 and outputs latent vector of 128, the input and output latent vector dimensions are denoted as: (128, 128). UGNN and RUGNN both have 4 graph levels, the latent vector feature lengths from graph level 0 to graph level 3 are 32, 32, 64, 128, respectively.

vanillaGNN and UGNN do not include a recurrent function in their processor blocks and therefore do not contain Edge processor GRU in processor. Since vanillaGNN and RGNN do not incorporate a downsample/upsample mechanism, they operate on the original finest graph level and require only a single processor block. In contrast, UGNN and the proposed RUGNN model include downsample/upsample mechanism, resulting in three processor blocks with specifications listed in the table.

**Appendix D Comparison of RUGNN predictions with FE ground truths for dome-shaped case study across timesteps**

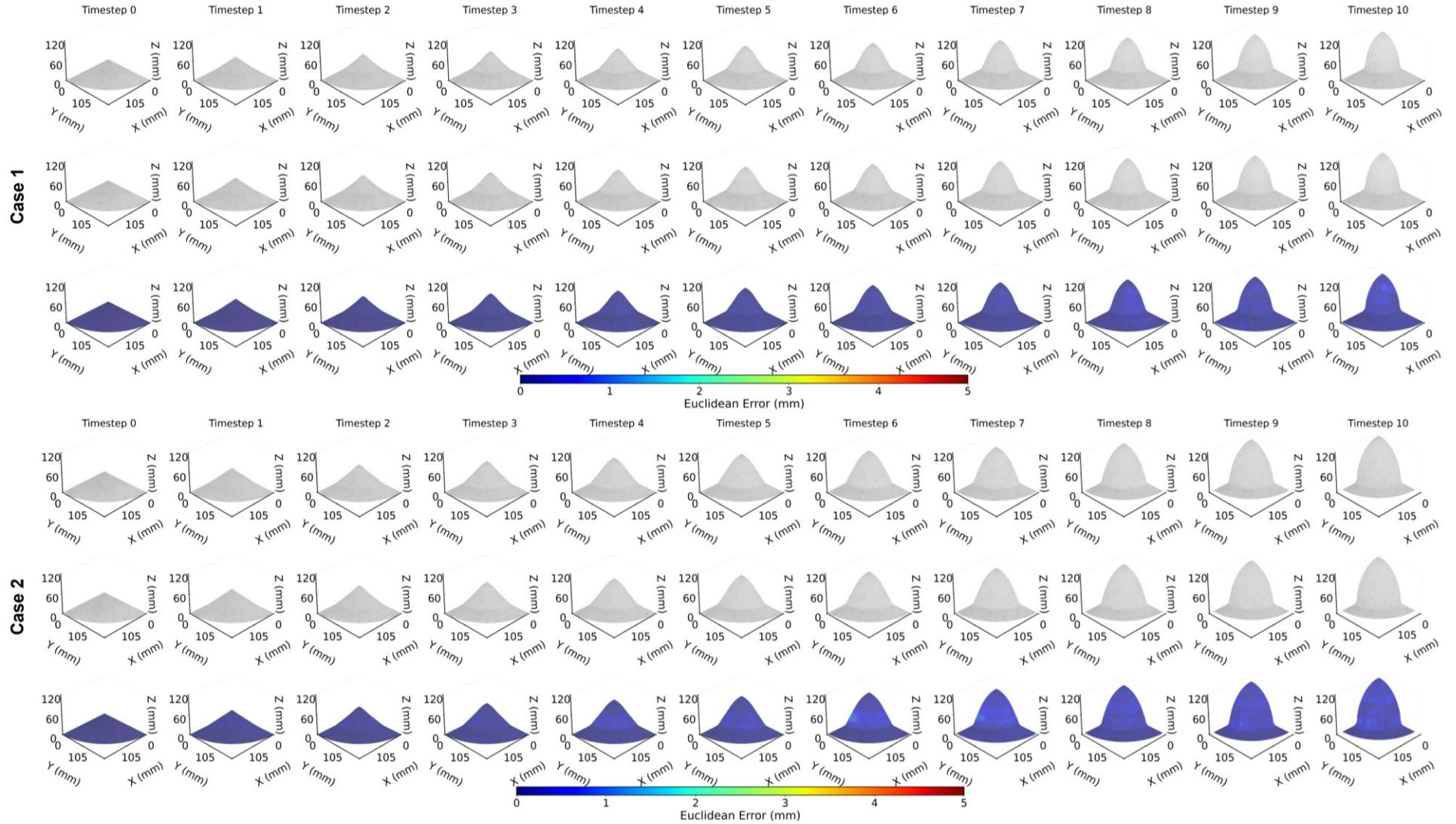

# Appendix E Model inputs and hyperparameters in bulkhead-shaped case study

Table E.1 Detailed comparison of model inputs and hyperparameters in bulkhead-shaped case study

| Model | Hyperparameters | | | Downsample/upsample |
|---|---|---|---|---|
| | **Processor block 1** | **Processor block 2** | **Processor block 3** | **Graph hierarchy** |
| **vanillaGNN** | Message passing layer = 10<br>Edge processor MLP (128, 128)<br>Node processor MLP (128, 128) | NA | NA | NA |
| **UGNN** | Message passing layer = 2<br>Edge processor MLP (32, 32)<br>Node processor MLP (32, 32) | Message passing layer = 10<br>Edge processor MLP (128, 128)<br>Node processor MLP (128, 128) | Message passing layer = 2<br>Edge processor MLP (32, 32)<br>Node processor MLP (32, 32) | 5 levels<br>(32, 32, 64, 128, 128) |
| **RUGNN** | (RGPB 1)<br>Message passing layer = 2<br>Edge processor MLP (32, 32)<br>Edge processor GRU (32, 32)<br>Node processor MLP (32, 32) | (RGPB 2)<br>Message passing layer = 10<br>Edge processor MLP (128, 128)<br>Edge processor GRU (128, 128)<br>Node processor MLP (128, 128) | (RGPB 3)<br>Message passing layer = 2<br>Edge processor MLP (32, 32)<br>Edge processor GRU (32, 32)<br>Node processor MLP (32, 32) | 5 levels<br>(32, 32, 64, 128, 128) |

Table E.1 summarises the key hyperparameters and architectural details of the baseline models and the proposed RUGNN. For vanillaGNN, the processor unit takes the latent space representation of 128 and outputs latent vector of 128, the input and output latent vector dimensions are denoted as: (128, 128). UGNN and RUGNN both have 5 graph levels, the latent vector feature lengths from graph level 0 to graph level 4 are 32, 32, 64, 128, 128, respectively.

vanillaGNN and UGNN do not include a recurrent function in their processor blocks and therefore lack hidden state inputs. Additionally, their processor blocks do not contain edge processor GRU layers. Since vanillaGNN does not incorporate a downsample/upsample mechanism, it operates on the original finest graph level and requires a single processor block. In contrast, UGNN and the proposed RUGNN model include downsample/upsample mechanism, resulting in three processor blocks with specifications listed in the table.

# Appendix F Comparison of RUGNN predictions with FE ground truths for bulkhead-shaped case study across timesteps

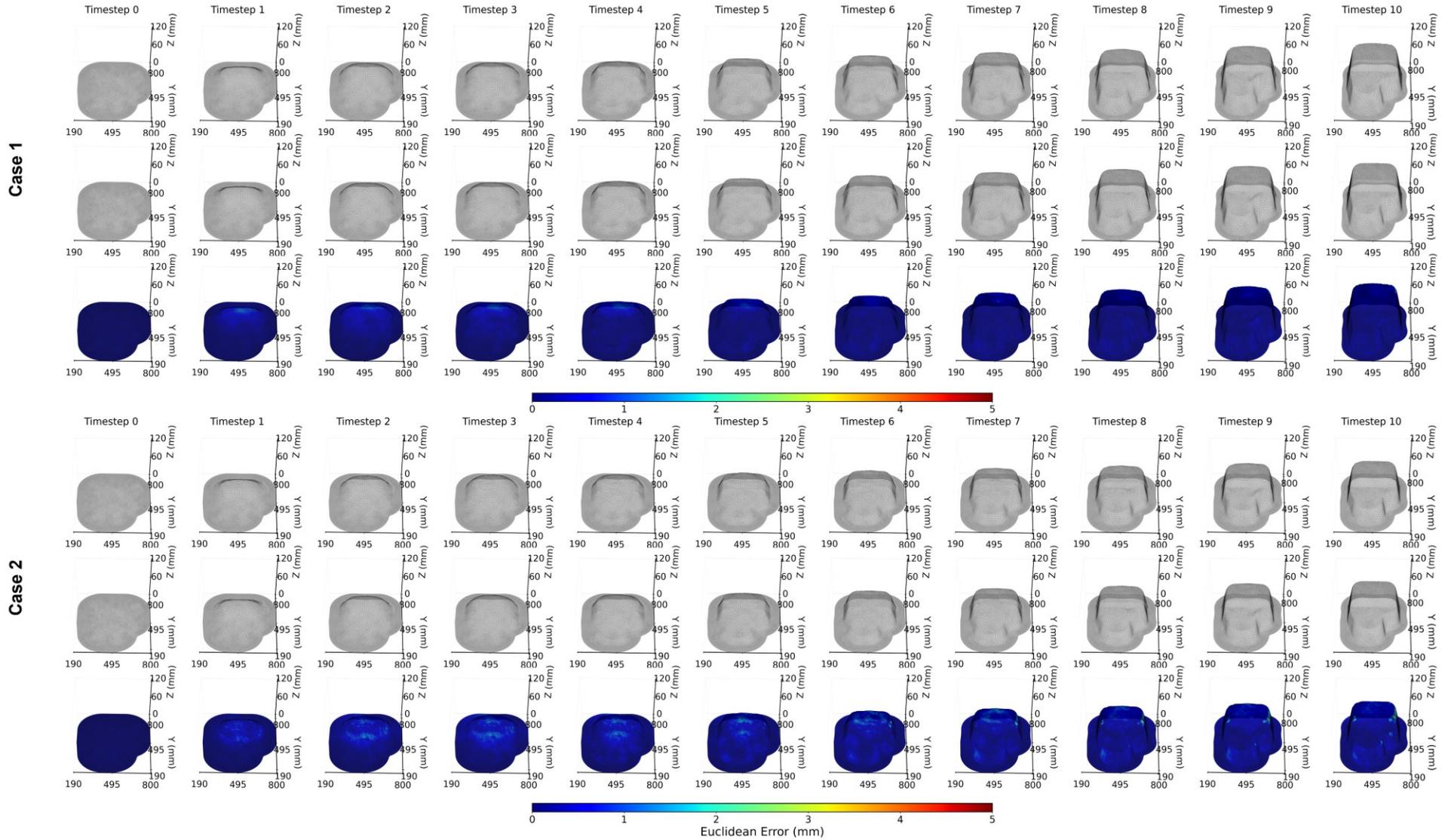

# References

[1]	J. L. D. Z. Marciniak, S.J. Hu, *Mechanics of Sheet Metal Forming*. Great Britain, 2002.

[2]	H. Attar, N. Li, and A. Foster, "A new design guideline development strategy for aluminium alloy corners formed through cold and hot stamping processes," *Materials & Design,* p. 109856, 05/01 2021, doi: 10.1016/j.matdes.2021.109856.

[3]	J. Zhou, B. Wang, J. Lin, and L. Fu, "Optimization of an aluminum alloy anti-collision side beam hot stamping process using a multi-objective genetic algorithm," *Archives of Civil and Mechanical Engineering,* vol. 13, no. 3, pp. 401-411, 2013/09/01/ 2013, doi: https://doi.org/10.1016/j.acme.2013.01.008.

[4]	K. Zheng, D. J. Politis, J. Lin, and T. A. Dean, "A study on the buckling behaviour of aluminium alloy sheet in deep drawing with macro-textured blankholder," *International Journal of Mechanical Sciences,* vol. 110, pp. 138-150, 2016/05/01/ 2016, doi: https://doi.org/10.1016/j.ijmecsci.2016.03.011.

[5]	H. Attar, H. Zhou, A. Foster, and N. Li, "Rapid feasibility assessment of components to be formed through hot stamping: A deep learning approach," *Journal of Manufacturing Processes,* vol. 68, pp. 1650-1671, 08/01 2021, doi: 10.1016/j.jmapro.2021.06.011.

[6]	H. Zhou, Q. Xu, Z. Nie, and N. Li, "A Study on Using Image-Based Machine Learning Methods to Develop Surrogate Models of Stamp Forming Simulations," *Journal of Manufacturing Science and Engineering,* vol. 144, pp. 1-41, 06/28 2021, doi: 10.1115/1.4051604.

[7]	L. Zhu and N. Li, "Springback prediction for sheet metal cold stamping using convolutional neural networks," in *2022 Workshop on Electronics Communication Engineering*, 2023, vol. 12720: SPIE, pp. 278-283.

[8]	S. Zhao, N. Wu, and Q. Wang, "Deep residual U-net with input of static structural responses for efficient U* load transfer path analysis," *Advanced Engineering Informatics,* vol. 46, p. 101184, 2020/10/01/ 2020, doi: https://doi.org/10.1016/j.aei.2020.101184.

[9]	J. Hou, C. Luo, F. Qin, Y. Shao, and X. Chen, "FuS-GCN: Efficient B-rep based graph convolutional networks for 3D-CAD model classification and retrieval," *Advanced Engineering Informatics,* vol. 56, p. 102008, 2023/04/01/ 2023, doi: https://doi.org/10.1016/j.aei.2023.102008.

[10]	J. Zhou *et al.*, "Graph neural networks: A review of methods and applications," *AI Open,* vol. 1, pp. 57-81, 2020/01/01/ 2020, doi: https://doi.org/10.1016/j.aiopen.2021.01.001.

[11]	T. Pfaff, M. Fortunato, A. Sanchez-Gonzalez, and P. W. Battaglia, "Learning mesh-based simulation with graph networks," *arXiv preprint arXiv:2010.03409,* 2020.

[12]	P. Veličković, "Everything is connected: Graph neural networks," *Current Opinion in Structural Biology,* vol. 79, p. 102538, 2023/04/01/ 2023, doi: https://doi.org/10.1016/j.sbi.2023.102538.

[13]	A. Sanchez-Gonzalez, J. Godwin, T. Pfaff, R. Ying, J. Leskovec, and P. W. Battaglia, "Learning to simulate complex physics with graph networks," presented at the Proceedings of the 37th International Conference on Machine Learning, 2020.

[14]	P. W. Battaglia *et al.*, "Relational inductive biases, deep learning, and graph networks," *arXiv preprint arXiv:1806.01261,* 2018.

[15]	Y. Zhao, H. Li, H. Zhou, H. R. Attar, T. Pfaff, and N. Li, "A review of graph neural network applications in mechanics-related domains," *Artificial Intelligence Review,* vol. 57, no. 11, p. 315, 2024/10/04 2024, doi: 10.1007/s10462-024-10931-y.

[16]	Y. Fei, S. Qin, W. Liao, H. Guan, and X. Lu, "Graph neural network-assisted evolutionary algorithm for rapid optimization design of shear-wall structures," *Advanced Engineering Informatics,* vol. 65, p. 103129, 2025/05/01/ 2025, doi: https://doi.org/10.1016/j.aei.2025.103129.

[17]	P. Zhao, Y. Fei, Y. Huang, Y. Feng, W. Liao, and X. Lu, "Design-condition-informed shear wall layout design based on graph neural networks," *Advanced Engineering Informatics,* vol. 58, p. 102190, 2023/10/01/ 2023, doi: https://doi.org/10.1016/j.aei.2023.102190.

[18]	N. Black and A. R. Najafi, "Learning finite element convergence with the Multi-fidelity Graph Neural Network," *Computer Methods in Applied Mechanics and Engineering,* vol. 397, 2022.

[19]	K. R. Allen *et al.*, "Learning rigid dynamics with face interaction graph networks," *arXiv preprint arXiv:2212.03574,* 2022.

[20]	M. Saleh, M. Sommersperger, N. Navab, and F. Tombari, "Physics-Encoded Graph Neural Networks for Deformation Prediction under Contact," *arXiv preprint arXiv:2402.03466,* 2024.

[21]	A. Sanchez-Gonzalez *et al.*, "Graph networks as learnable physics engines for inference and control," in *International Conference on Machine Learning*, 2018: PMLR, pp. 4470-4479.

[22]	M. Lino, S. Fotiadis, A. A. Bharath, and C. D. Cantwell, "Towards Fast Simulation of Environmental Fluid Mechanics with Multi-Scale Graph Neural Networks," *ArXiv,* vol. abs/2205.02637, 2022.

[23]	Q. Chen, J. Cao, W. Lin, S. Zhu, and S. Wang, "Predicting dynamic responses of continuous deformable bodies:A graph-based learning approach," *Computer Methods in Applied Mechanics and Engineering,* vol. 420, p. 116669, 2024/02/15/ 2024, doi: https://doi.org/10.1016/j.cma.2023.116669.

[24]	H. Gao, M. J. Zahr, and J.-X. Wang, "Physics-informed graph neural Galerkin networks: A unified framework for solving PDE-governed forward and inverse problems," *Computer Methods in Applied Mechanics and Engineering,* vol. 390, p. 114502, 2022/02/15/ 2022, doi: https://doi.org/10.1016/j.cma.2021.114502.

[25]	D. Dalton, D. Husmeier, and H. Gao, "Physics-informed graph neural network emulation of soft-tissue mechanics," *Computer Methods in Applied Mechanics and Engineering,* vol. 417, p. 116351, 2023/12/01/ 2023, doi: https://doi.org/10.1016/j.cma.2023.116351.

[26]	J. He, D. Abueidda, S. Koric, and I. Jasiuk, "On the use of graph neural networks and shape-function-based gradient computation in the deep energy method," *International Journal for Numerical Methods in Engineerin,* vol. 124, 2023.

[27]	J. Jeswiet *et al.*, "Metal forming progress since 2000," *CIRP Journal of Manufacturing Science and Technology,* vol. 1, no. 1, pp. 2-17, 2008/01/01/ 2008, doi: https://doi.org/10.1016/j.cirpj.2008.06.005.

[28]  D. M. Neto, J. Coër, M. C. Oliveira, J. L. Alves, P. Y. Manach, and L. F. Menezes, "Numerical analysis on the elastic deformation of the tools in sheet metal forming processes," *International Journal of Solids and Structures,* vol. 100-101, pp. 270-285, 2016/12/01/ 2016, doi: https://doi.org/10.1016/j.ijsolstr.2016.08.023.

[29]  N. Li *et al.*, "Experimental investigation of boron steel at hot stamping conditions," *Journal of Materials Processing Technology,* vol. 228, pp. 2-10, 2016/02/01/ 2016, doi: https://doi.org/10.1016/j.jmatprotec.2015.09.043.

[30]  M. S. Mohamed, A. D. Foster, J. Lin, D. S. Balint, and T. A. Dean, "Investigation of deformation and failure features in hot stamping of AA6082: Experimentation and modelling," *International Journal of Machine Tools and Manufacture,* vol. 53, no. 1, pp. 27-38, 2012/02/01/ 2012, doi: https://doi.org/10.1016/j.ijmachtools.2011.07.005.

[31]  S. Deshpande, S. P. A. Bordas, and J. Lengiewicz, "MAgNET: A graph U-Net architecture for mesh-based simulations," *Engineering Applications of Artificial Intelligence,* vol. 133, p. 108055, 2024/07/01/ 2024, doi: https://doi.org/10.1016/j.engappai.2024.108055.

[32]  M. Fortunato, T. Pfaff, P. Wirnsberger, A. Pritzel, and P. W. Battaglia, "MultiScale MeshGraphNets," *ArXiv,* vol. abs/2210.00612, 2022.

[33]  Y. Rubanova, T. Lopez-Guevara, K. R. Allen, W. F. Whitney, K. Stachenfeld, and T. Pfaff, "Learning rigid-body simulators over implicit shapes for large-scale scenes and vision," *arXiv preprint arXiv:2405.14045,* 2024.

[34]  B. Yu, H. Yin, and Z. Zhu, "St-unet: A spatio-temporal u-network for graph-structured time series modeling," *arXiv preprint arXiv:1903.05631,* 2019.

[35]  B. Liang, S. Gao, and W. Zhang, "An integrated preforming-performance model for high-fidelity performance analysis of cured woven composite part with non-orthogonal yarn angles," *Chinese Journal of Aeronautics,* vol. 35, no. 6, pp. 367-378, 2022/06/01/ 2022, doi: https://doi.org/10.1016/j.cja.2021.09.019.

[36]  M. Mohamed *et al.*, "An investigation of a new 2D CDM model in predicting failure in HFQing of an automotive panel," *MATEC Web of Conferences,* vol. 21, p. 05011, 2015. [Online]. Available: https://doi.org/10.1051/matecconf/20152105011.

[37]  A. Foster, M. Mohamed, J. Lin, and T. Dean, "An investigation of lubrication and heat transfer for a sheet aluminium heat, form-quench (HFQ) process," *Steel Research International,* vol. 79, pp. 113-120, 02/01 2008.

[38]  J. O. Hallquist, G. L. Goudreau, and D. J. Benson, "Sliding interfaces with contact-impact in large-scale Lagrangian computations," *Computer Methods in Applied Mechanics and Engineering,* vol. 51, no. 1, pp. 107-137, 1985/09/01/ 1985, doi: https://doi.org/10.1016/0045-7825(85)90030-1.

[39]  J. Gilmer, S. S. Schoenholz, P. F. Riley, O. Vinyals, and G. E. Dahl, "Neural message passing for quantum chemistry," in *International conference on machine learning*, 2017: PMLR, pp. 1263-1272.

[40]  Y. Cao, M. Chai, M. Li, and C. Jiang, "Efficient learning of mesh-based physical simulation with bi-stride multi-scale graph neural network," presented at the Proceedings of the 40th International Conference on Machine Learning, Honolulu, Hawaii, USA, 2023.